\documentclass[10pt,twocolumn,letterpaper]{article}

\usepackage{iccv}
\usepackage{times}
\usepackage{epsfig}
\usepackage{graphicx}
\usepackage{amsmath}
\usepackage{amssymb}
\usepackage[accsupp]{axessibility}

\usepackage{microtype}
\usepackage{booktabs} %
\usepackage[numbers]{natbib}

\usepackage[pagebackref=true,breaklinks=true,letterpaper=true,colorlinks,bookmarks=false]{hyperref}

\iccvfinalcopy %
\def\iccvPaperID{5806} %

\ificcvfinal\fi

\usepackage{amsmath}
\usepackage{amssymb}
\usepackage{mathtools}
\usepackage{amsthm}

\usepackage[capitalize,noabbrev]{cleveref}

\theoremstyle{plain}

\theoremstyle{definition}

\theoremstyle{remark}

\Crefname{equation}{Eq.}{Eqs.}
\Crefname{figure}{Fig.}{Figs.}
\Crefname{tabular}{Tab.}{Tabs.}

\usepackage{amsmath,amsfonts,bm}

\def\eqref#1{equation~\ref{#1}}

\def\1{\bm{1}}

\def\mM{{\bm{M}}}

\def\mP{{\bm{P}}}

\def\mX{{\bm{X}}}

\DeclareMathAlphabet{\mathsfit}{\encodingdefault}{\sfdefault}{m}{sl}
\SetMathAlphabet{\mathsfit}{bold}{\encodingdefault}{\sfdefault}{bx}{n}
\newcommand{\tens}[1]{\bm{\mathsfit{#1}}}

\def\tP{{\tens{P}}}

\newcommand{\R}{\mathbb{R}}

\DeclareMathOperator*{\argmin}{arg\,min}

\usepackage{amsfonts}
\usepackage[utf8]{inputenc} %
\usepackage[T1]{fontenc}    %
\usepackage{nicefrac}       %
\usepackage{microtype}      %
\usepackage{multirow}
\usepackage{makecell}
\usepackage{courier}
\usepackage{xspace}
\usepackage{bm}
\usepackage{wrapfig}
\usepackage{oubraces}
\usepackage{amsthm}
\usepackage{caption}
\usepackage{subcaption}
\usepackage{tabularx}
\usepackage{enumitem}
\usepackage{stfloats}
\usepackage{url}
\usepackage{enumitem}
\usepackage{pythonhighlight}

\newcolumntype{Y}{>{\centering\arraybackslash}X}

\def\eqref#1{Eqn.~\ref{#1}}

\def\inch#1{#1''}

\newcommand{\chawin}[1]{}
\newcommand{\note}[1]{}
\newcommand{\todo}[1]{}
\newcommand{\future}[1]{}
\newcommand{\mylink}[1]{\url{#1}}

\newcommand{\psize}[2]{\inch{#1}$\times$\inch{#2}}
\newcommand{\apb}{REAP\xspace}
\newcommand{\apbs}{REAP-S\xspace}

\begin{document}

\title{REAP: A Large-Scale Realistic Adversarial Patch Benchmark}

\author{
Nabeel Hingun\textsuperscript{*}\\
UC Berkeley\\
{\tt\small nabeel126@berkeley.edu}
\and
Chawin Sitawarin\textsuperscript{*}\\
UC Berkeley\\
{\tt\small chawins@berkeley.edu}
\and
Jerry Li\\
Microsoft\\
{\tt\small jerrl@microsoft.com}
\and
David Wagner\\
UC Berkeley\\
{\tt\small daw@cs.berkeley.edu}
}

\ificcvfinal\thispagestyle{empty}\fi

\twocolumn[{%
                  \renewcommand\twocolumn[1][]{#1}%
                  \maketitle
                  \begin{center}
                        \vspace{-20pt}
                        \centering
                        \captionsetup{type=figure}
                        \includegraphics[width=0.85\linewidth]{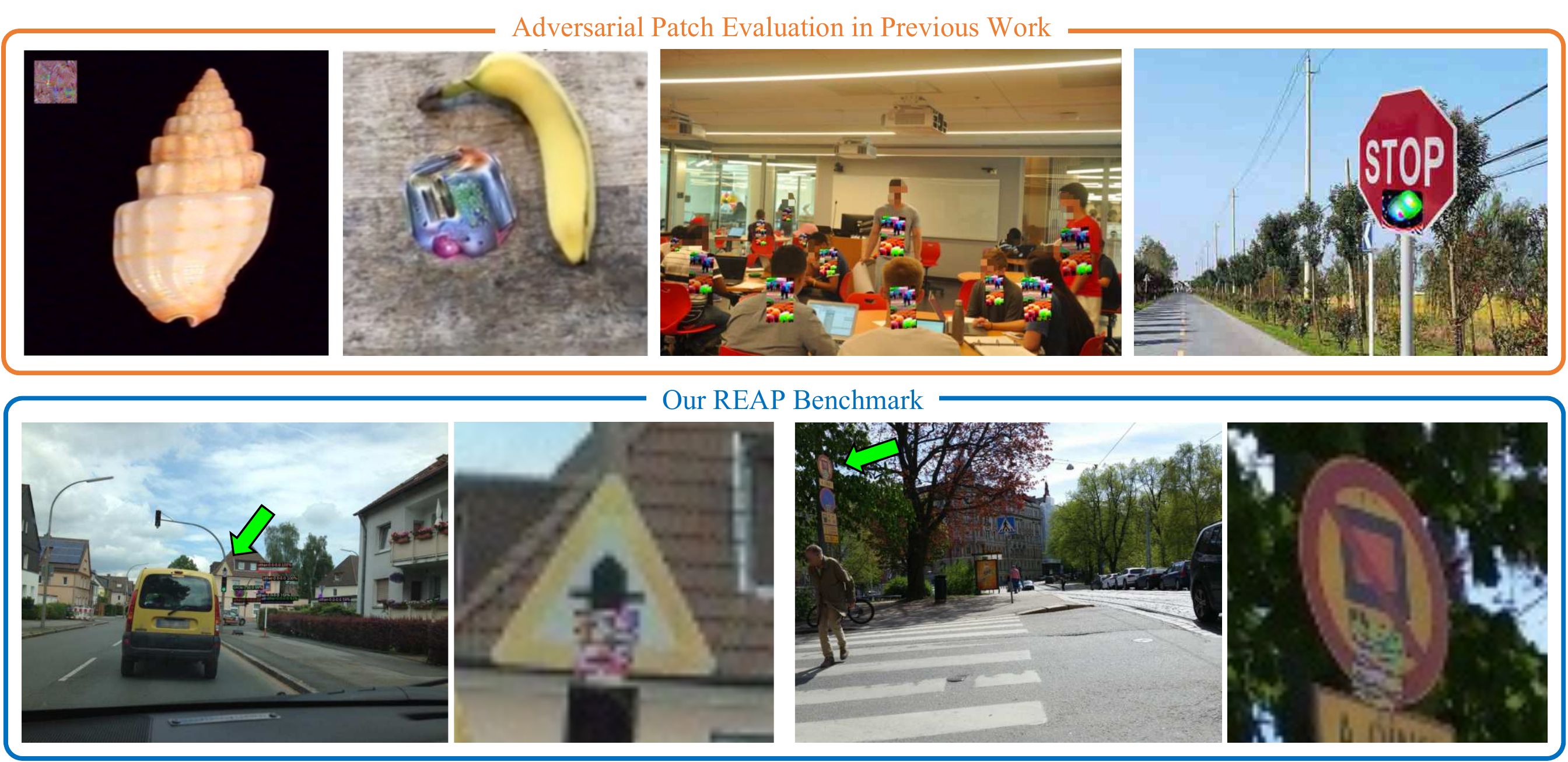}
                        \captionof{figure}{
                              When digitally evaluating patch attacks, past work (top row) ignores many real-world factors and thus may yield a misleading evaluation. We develop the \apb{} benchmark (bottom row) that more realistically simulates the effect of a real-world patch attack on road signs, accounting for the pose, the location, and the lighting condition.
                        }\label{fig:banner}
                  \end{center}%
            }]

\begin{abstract}
      Machine learning models are known to be susceptible to adversarial perturbation.
      One famous attack is the \emph{adversarial patch}, a particularly crafted sticker that makes the model mispredict the object it is placed on.
      This attack presents a critical threat to cyber-physical systems that rely on cameras such as autonomous cars.
      Despite the significance of the problem, conducting research in this setting has been difficult;
      evaluating attacks and defenses in the real world is exceptionally costly while synthetic data are unrealistic.
      In this work, we propose the \apb{} (REalistic Adversarial Patch) benchmark, a digital benchmark that enables the evaluations on real images under real-world conditions.
      Built on top of the Mapillary Vistas dataset, our benchmark contains over 14,000 traffic signs.
      Each sign is augmented with geometric and lighting transformations for applying a digitally generated patch realistically onto the sign.
      Using our benchmark, we perform the first large-scale assessments of adversarial patch attacks under realistic conditions.
      Our experiments suggest that patch attacks may present a smaller threat than previously believed and that the success rate of an attack on simpler digital simulations is not predictive of its actual effectiveness in practice.
      Our benchmark is released publicly
      at \mylink{https://github.com/wagner-group/reap-benchmark}
\end{abstract}

\vspace{-15pt}
\section{Introduction}

\let\thefootnote\relax\footnotetext{\textsuperscript{*}Equal contribution.}Research has shown that machine learning models lack robustness against adversarially chosen perturbations.
\citet{szegedy_intriguing_2014} first demonstrated that one can engineer perturbations that are indiscernible to the human eye yet that cause neural networks to misclassify images with high confidence.
Since then, there has been a large body of academic work on understanding the robustness of neural networks to such attacks~\citep{goodfellow_explaining_2015,moosavi-dezfooli_deepfool_2016,tanay_boundary_2016,carlini_evaluating_2017,kurakin_adversarial_2017,tramer_space_2017,madry_deep_2018,bubeck_adversarial_2019,ilyas_adversarial_2019}.

One particularly concerning type of attack is the \emph{adversarial patch attack}~\citep{brown_adversarial_2018,eykholt_physical_2018,karmon_lavan_2018,sitawarin_darts_2018,chen_shapeshifter_2019,jan_connecting_2019,liu_dpatch_2019,patel_adaptive_2019,sharma_attacks_2019,zhao_seeing_2019,huang_universal_2020,wu_making_2020}.
These are real-world attacks, where the attacker's objective is to print out a patch, physically place it in a scene, and cause a vision network processing the scene to malfunction.
These attacks are especially concerning because of the potential impact on autonomous vehicles.
A malicious agent could, for instance, produce a sticker that, when placed on a stop sign, cause a self-driving car to believe it is (say) a speed limit sign, and fail to stop.
Indeed, similar attacks have already been demonstrated both in academic settings~\citep{liu2019patch,evtimov_robust_2017,sato2021dirty} and on real-world autonomous vehicles~\citep{tencent2019}.

Despite this significant risk, research on these attacks has stalled to a certain extent because quantitatively evaluating the significance of this threat is challenging.
The most accurate approach would be to conduct real-world experiments, but they are very expensive, and at present, not practical to do at a large scale.
This leaves much to be desired compared to other branches of computer vision research, where the availability of benchmarks such as ImageNet have reduced the barriers to research and spurred tremendous innovation.

Instead, researchers turn to one of two techniques: either, they physically create their attacks and try them out on a small number of real-world examples by physically attaching them to objects, or they digitally evaluate patch attacks using digital images containing simulated patches.
Both approaches have major drawbacks.
Although the former simulates more realistic conditions, the sample size is very small, and typically one cannot draw statistical conclusions from the results~\citep{brown_adversarial_2018,eykholt_physical_2018,sitawarin_darts_2018,chen_shapeshifter_2019,zhao_seeing_2019,hoory_dynamic_2020,huang_universal_2020,wu_making_2020}.
Additionally, because of the ad-hoc nature of these evaluations, it is impossible to compare the results across different papers.
Ultimately, such experiments only serve as a proof of concept for the proposed attacks and defenses, but not as a rigorous evaluation of their effectiveness.

In contrast, a digital simulation of attacks/defenses allows quantitative evaluation~\citep{karmon_lavan_2018,liu_dpatch_2019,zhang_clipped_2020,rao_adversarial_2020,mccoyd_minority_2020,xiang_patchguard_2021,wang_physicalworld_2021,pintor_imagenetpatch_2023}.
However, it is difficult to accurately capture all of the challenges that arise in the real world.
Past work often made unrealistic assumptions, such as that the patch is square, axis-aligned, can be located anywhere on the image, fully under the control of the attacker, and ignore noise and variation in lighting and pose (see top row of \cref{fig:banner}).
Consequently, it is unclear if these evaluations are actually reflective of what would happen in real-world scenarios.

\subsection{Our Contributions}\vspace{-2pt}

\noindent \textbf{The \apb Benchmark:}
We propose REalistic Adversarial Patch Benchmark (\apb), the first large-scale standardized benchmark for security against patch attacks.
Motivated by the aforementioned shortcomings of prior evaluations, we design \apb with the following principles in mind:
\begin{enumerate}[leftmargin=*,nosep]
      \item {\bf Large-scale evaluation:} \apb consists of 14,651 images of road signs drawn from the Mapillary Vistas dataset.
            This allows us to draw quantitative conclusions about the effectiveness of attacks/defenses on the dataset.
      \item {\bf Realistic patch rendering:} \apb has tooling, which, for every road sign in the dataset, allows us to realistically render any digital patch onto the sign, matching factors such as where to place the patch, the camera angle, lighting conditions, etc.
            Importantly, this transformation is fast and differentiable so one can still perform backpropagation through the rendering process.
      \item {\bf Realistic image distribution:} \apb consists of images taken under realistic conditions, including variation in sizes and distances from the camera as well as various lighting conditions and degrees of occlusion.
\end{enumerate}

\smallskip \noindent \textbf{Evaluations with \apb:}
With our new benchmark in hand, we also perform the first large-scale evaluations of existing attacks on object detectors.
We evaluate existing attacks on three different object detection architectures: Faster R-CNN~\citep{ren_faster_2015}, YOLOF~\citep{chen_you_2021}, and DINO~\citep{zhang_dino_2022}.
We also implement and evaluate a baseline defense adapted from adversarial training~\citep{madry_deep_2018} to defend against patch attacks on object detection.
The conclusions we find are:
\begin{enumerate}[leftmargin=*,nosep]
      \item {\bf Existing patch attacks are not that effective.} Perhaps surprisingly, existing attacks do not succeed on a majority of images on our benchmark.
            This is in contrast to simpler attack models such as $\ell_p$-bounded perturbations or patch attacks on simpler benchmarks, where the attack success rate is near $100\%$.
            Moreover, adversarially trained models can almost completely stop the attacks with only a minor performance drop on benign data.
      \item {\bf Performance on synthetic data is not reflective of performance on \apb.} We find that the success rates of attacks on synthetic versions of our benchmark and the full \apb are only poorly correlated.
            We conclude that performance on simple synthetic benchmarks is not predictive of attack success rate in more realistic conditions.
      \item {\bf Lighting and patch placement are particularly important.} Finally, we investigate what transforms in the patch rendering are the most important, in terms of the effect on the attack success rate.
            We find that the most significant first-order effects are from the lighting transform, as well as the positioning of the patch.
            In contrast, the perspective transforms---while still important---seem to affect the attack success rate somewhat less.
\end{enumerate}
While we believe these conclusions are already quite interesting, they are only the tip of the iceberg of what can be done with \apb.
We believe that \apb will help support future research in adversarial patches by enabling a more accurate evaluation of new attacks and defenses.

\section{Related Work}\vspace{-2pt}

\noindent \textbf{Adversarial patch attacks.} The literature on adversarial patches, and adversarial attacks more generally, is vast and a full review is beyond the scope of this paper.
For conciseness, we only survey the most relevant works.
Since their introduction in \citet{brown_adversarial_2018,karmon_lavan_2018,eykholt_physical_2018}, there have been a variety of adversarial patch attacks proposed~\citep{jan_connecting_2019,liu_dpatch_2019,patel_adaptive_2019,sharma_attacks_2019,huang_universal_2020,wu_making_2020}.
Of particular interest to us are the ones on object detection of road signs~\citep{eykholt_physical_2018,sitawarin_darts_2018,chen_shapeshifter_2019,zhao_seeing_2019}.

\smallskip \noindent \textbf{Small scale, real-world tests.}
A common methodology used to test the transferability of the adversarial patch to the physical world is to print it out, physically place it onto an object, and capture pictures or videos of the patch for evaluation~\citep{brown_adversarial_2018,eykholt_physical_2018,sitawarin_darts_2018,chen_shapeshifter_2019,zhao_seeing_2019,hoory_dynamic_2020,huang_universal_2020,wu_making_2020}.
While this method provides the most realistic evaluation, it has a number of downsides.
First, it is, by nature, very time-consuming and hence limits the number of images that can be used for testing.
Consequently, one cannot extract quantitative conclusions from the results.
Additionally, they are difficult to standardize across papers, making their result not directly comparable.
For instance, the pictures of the adversarial patches are taken under different angles, lighting conditions, or from varying distances.
Sometimes, the adversarial patches themselves are printed using different printers~\citep{chen_shapeshifter_2019,zhao_seeing_2019}.

\smallskip \noindent \textbf{Completely simulated environment.}
Another line of work considers purely simulated environments for evaluating adversarial patches such as CARLA~\citep{dosovitskiy_carla_2017,liu_synthetic_2022,ramakrishna_anticarla_2022} and AttackScenes~\citep{huang_universal_2020}.
A huge advantage of this method is that it has the most precise and the most flexible control of the environment, e.g., cameras and objects can be placed anywhere.
However, it is labor-intensive to build a diverse set of scenes digitally, and it compromises heavily on realism.
Another example is 3DB~\citep{leclerc20213db}, a photorealistic simulation for studying the reliability of computer vision systems.
Nevertheless, it lacks the tooling necessary for evaluating adversarial patches and does not contain any driving scene, a setting to which adversarial patches are most applicable.
Our benchmark utilizes images of real and diverse driving scenes and focuses on realistically simulating only the adversarial patches.

\smallskip \noindent \textbf{Digital simulation.}
This third approach takes a middle road and simulates the effects of the adversarial patch by digitally inserting it into a real image.
This has been done at scale and to varying degrees of sophistication.
One of the most common, but also simplest, ways this is done is to apply the patch to the image at some random position, and with some simple transformations, for instance, those induced by expectation over transformation~\citep{brown_adversarial_2018,karmon_lavan_2018,liu_dpatch_2019,zhang_clipped_2020,rao_adversarial_2020,mccoyd_minority_2020,xiang_patchguard_2021,wang_physicalworld_2021,pintor_imagenetpatch_2023}.
This approach violates all the physical constraints and hence, is far from being realistic.

Arguably the benchmarks most similar to ours are the ones in \citet{zhao_seeing_2019} and \citet{braunegg_apricot_2020}.
\citet{zhao_seeing_2019} digitally insert synthetic stop signs with patches into images with realistic camera angles.
However, they do not account for lighting conditions, and the target object itself is synthetic.
In contrast, all signs in our dataset are real, and we also produce a transformation to match lighting conditions.
In \cref{ssec:results}, we find that these two factors affect the evaluation metrics to a large extent.
APRICOT~\citep{braunegg_apricot_2020} contains images of real scenes with a printed adversarial patch.
Compared to ours, APRICOT is smaller in size (1K vs 14K images) and is heavily inflexible as it comes with a pre-defined adversarial patch with a fixed size and location.

\smallskip \noindent \textbf{Defenses.}
There have also been a slew of proposed defenses against patch attacks,  e.g.,~\citep{hayes_visible_2018,naseer_local_2019,zhang_clipped_2020,xiang_patchguard_2021,rao_adversarial_2020,mccoyd_minority_2020,mu2021rsa,chen_turning_2021}.
Most examine object classification.
Only a handful consider object detection, which may be more relevant in practice~\citep{chou_sentinet_2020,xiang_objectseeker_2023}.
We choose to experiment with adversarial training~\citep{madry_deep_2018} as a defense because, to the extent of our knowledge, it has not been applied in this setting (\citet{rao_adversarial_2020} study patch adversarial training on classifiers).
It is also known to be a strong baseline and arguably the only effective defense across other $\ell_p$-norms~\citep{croce_robustbench_2020}.
Importantly, unlike the other defenses listed above, adversarial training does not make assumptions about the number or size of the patch.

\section{Adversarial Patch Benchmark}\vspace{-2pt}

\subsection{Overview}\vspace{-2pt}

Our dataset is a collection of images containing traffic signs, each of which comes with a segmentation mask and a class.
So far, this is more or less standard.
The main additional feature of our benchmark is that, for each sign, we also provide an associated rendering transformation.

Given a digital patch, this transformation allows us to apply the patch on the sign in a way that respects the scaling, orientation, and lighting of the sign in the image.
We emphasize that a separate transformation is inferred individually for each sign, in order to ensure that the transformation is accurate for every image.
Moreover, the rendering transformation is fully differentiable, which allows our dataset to be used to generate patch attacks and to apply adversarial defenses along the line of adversarial training.

\cref{fig:compute_geo_tf_steps,fig:compute_light_tf_steps} give an overview of the process to obtain the \emph{geometric} (\cref{sssec:geo}) and the \emph{relighting} (\cref{sssec:light}) transformations, respectively.
We use an algorithm to generate the candidate annotations automatically, visually inspect each of them, and then manually fix any wrong annotation.
In total, we label 14,651 traffic signs across 8,433 images.

\subsection{Datasets}

We build our benchmark using images from the Mapillary Vistas dataset~\citep{neuhold_mapillary_2017}.
It includes 20,000 street-level images from around the world, annotated with bounding boxes of 124 object categories, including traffic signs.
A limitation of Vistas is that all traffic signs are grouped under one class.
This creates a challenge for us, because our patch-rendering process depends on the size and shape of the sign.
Without this information, the rendering is less realistic.
We deal with this challenge by grouping the signs so that all signs in the same group can use the same geometric transform procedure.
The grouping process is described in the next section.

\subsection{Traffic Sign Classification}\label{ssec:classify}

Grouping traffic signs by their shape and size has two advantages.
First, it allows more accurate geometric transforms as previously mentioned.
Second, it allows us to study multi-class sign detection.
Instead of labeling the Vistas signs by hand, we train a ResNet-18 on a similar dataset, Mapillary Traffic Sign (MTSD)~\citep{ertler_mapillary_2020}, to classify them.
MTSD contains granular labeling of over 300 traffic sign classes, but we cannot use it in place of Vistas as it lacks segmentation labels required to compute the geometric transforms.

\textbf{We created two versions of the benchmark: \apb{} and \apbs{}.}\footnote{Update \today: In the previous version of the paper, we only present \apbs{} which was called \apb{}.}
\apb{} is our main benchmark with the classes matching those of MTSD.
However, most of the classes contain fewer than 10 samples so we only keep the 100 most common classes.
We need a sufficient number of samples per class because (i) some will be further filtered out in the preprocessing and (ii) the samples will be split into a ``training'' (for the attacker to ``train'' the adversarial patch) and a test set (for evaluating the attack).
Conversely, \apbs{} groups the signs into 11 classes by shape and size, namely circle, triangle, upside-down triangle, diamond (S), diamond (L), square, rectangle (S), rectangle (M), rectangle (L), pentagon, and octagon.
\apbs{} serves as a simpler alternative to \apb{} and is also intuitively more ``defender-friendly.''
For both \apb{} and \apbs{}, the remaining signs that do not belong to the classes are labeled as a background class or ``others'' which will be ignored when we compute the metrics.

Since Mapillary Vistas does not come with these labels, we first trained a ConvNeXt model~\citep{liu_convnet_2022} on MTSD, which achieves about 98\% accuracy on the validation set, to generate the candidate class labels.
The labels were then automatically corrected when we compute the parameters for the geometric transform in \cref{sssec:geo}.
For the remaining ones that cannot be automatically verified, we manually inspected and corrected them.

Our grouping of the signs in \apbs{} has an extra benefit.
Since each class is (approximately) associated with a standardized physical size, we can specify the patch size in real units (e.g., inches) instead of pixels.
The real unit is arguably more useful for estimating the threat of adversarial patches than constraining the size by the number of pixels.
One complication is that a single class of sign may come in different sizes, e.g., stop signs can be \inch{24}, \inch{36}, or \inch{48}, depending on the kind of road they are located on, but usually one size is more common.
The Vistas dataset does not contain sufficient information to distinguish between these sizes so we pick one canonical size for each sign type.
Specifically, we select the size specified for ``Expressway'' according to the official U.S. Department of Transportation's guideline.\footnote{\url{https://mutcd.fhwa.dot.gov/htm/2003/part2/part2b1.htm}}
\cref{ap:sec:anno_detail} describes our design decision in detail.

\begin{figure*}[t]
    \centering
    \includegraphics[width=0.9\textwidth]{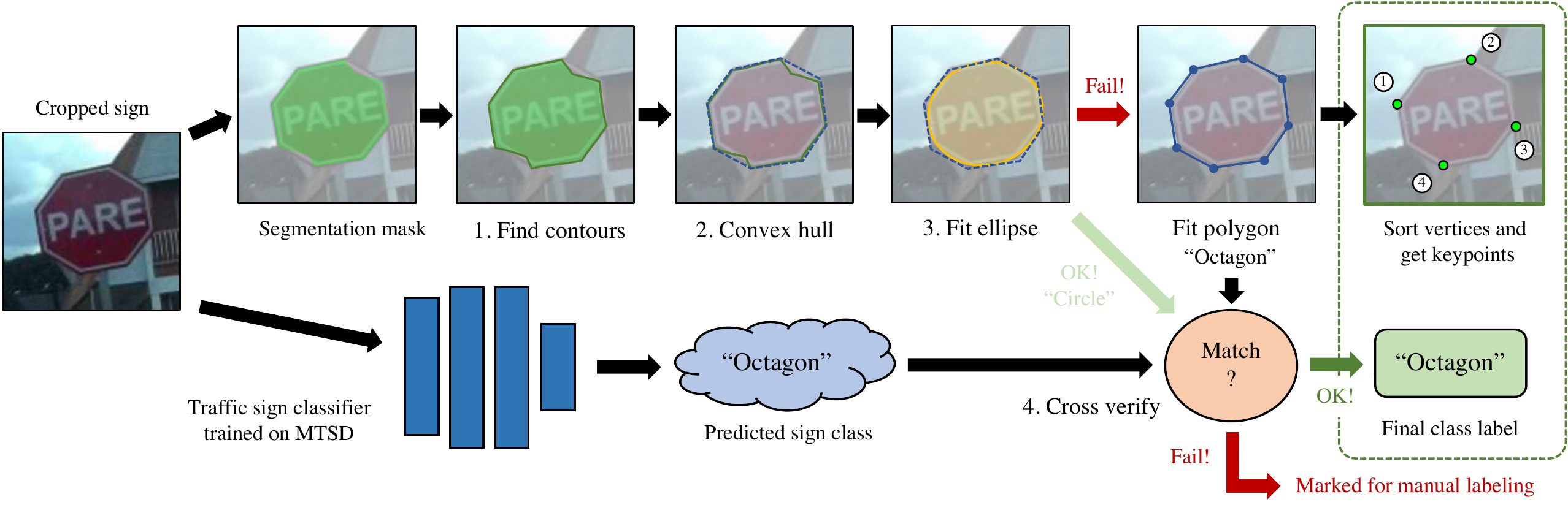}
    \vspace{-10pt}
    \caption{The automated procedure we use to extract the keypoints from each traffic sign.}\label{fig:compute_geo_tf_steps}
\end{figure*}

\subsection{Transformations}\label{ssec:transform}

\begin{figure}[t]
    \centering
    \includegraphics[width=\linewidth]{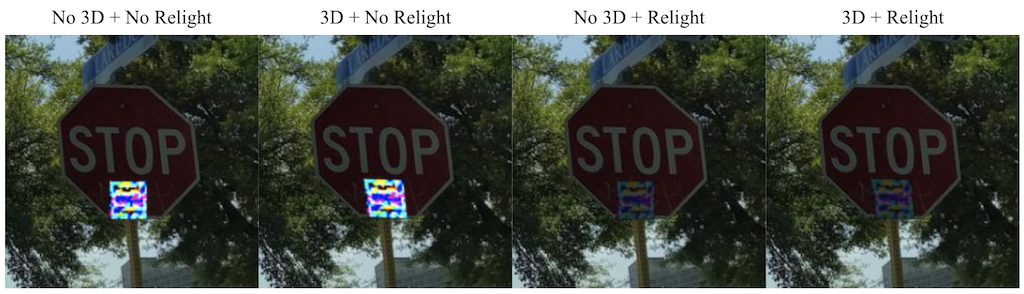}
    \vspace{-15pt}
    \caption{Example ablation of the geometric and relighting transforms in our dataset. The rightmost stop sign has a patch rendered with a perspective and relighting transform which makes it more realistic. The first and second images have patches that are too bright whereas the first and third images have patches that do not respect the sign's orientation.}\label{fig:transform_ablation}
\end{figure}

We render adversarial patches with two types of transformations: \emph{geometric} and \emph{relighting}.
Since the traffic signs in our dataset vary in shape, size, and orientation, we first need to apply a geometric, specifically perspective or 3D, transform to the patch to simulate these variations.
Next, we account for the fact that pictures of real-world traffic signs are taken under different lighting conditions by applying a relighting transform to the patch.
The importance of these transformations is highlighted in \cref{fig:transform_ablation}.

\subsubsection{Geometric Transformation}\label{sssec:geo}

To determine the parameters of the perspective transform, we need four keypoints for each sign.
We infer the keypoints for a particular traffic sign using only its segmentation mask (which is provided in the Mapillary Vistas dataset) by following the four steps below (also visualized in \cref{fig:compute_geo_tf_steps}):

\begin{enumerate}[leftmargin=*,nosep]
    \item \textbf{Find contour}: First, we find the contour of the segmentation mask.
    \item \textbf{Compute convex hull}: Then, we find the convex hull of the contour to correct annotation errors and occlusion. This does not affect correct masks, as they should already be convex.
    \item \textbf{Fit polygon and ellipse}: We fit an ellipse to the convex hull, to find circular signs. If the fitted ellipse results in a larger error than a certain threshold, we know that the sign is not circular and therefore fit a polygon instead.
    \item \textbf{Cross verify}: We verify that the shape obtained from the previous step matches with the ResNet's prediction. If not, the sign is flagged for manual inspection.
\end{enumerate}

The last step is finding the keypoints.
For polygons, we first match the vertices to the canonical ones and then take the four predefined vertices as the keypoints.
For circular signs, we use the ends of their major and minor axes as the four keypoints.
These keypoints are used to infer a perspective transform appropriate for this sign.
Triangular signs are a special case as we can only identify a maximum of three keypoints which means we can only infer a unique affine transform (six degrees of freedom).
Note that this transform is linear and hence is fully differentiable.
Lastly, we manually check all annotations and correct any errors.

\subsubsection{Relighting Transformation}\label{sssec:light}

\begin{figure}[t]
    \centering
    \includegraphics[width=\linewidth]{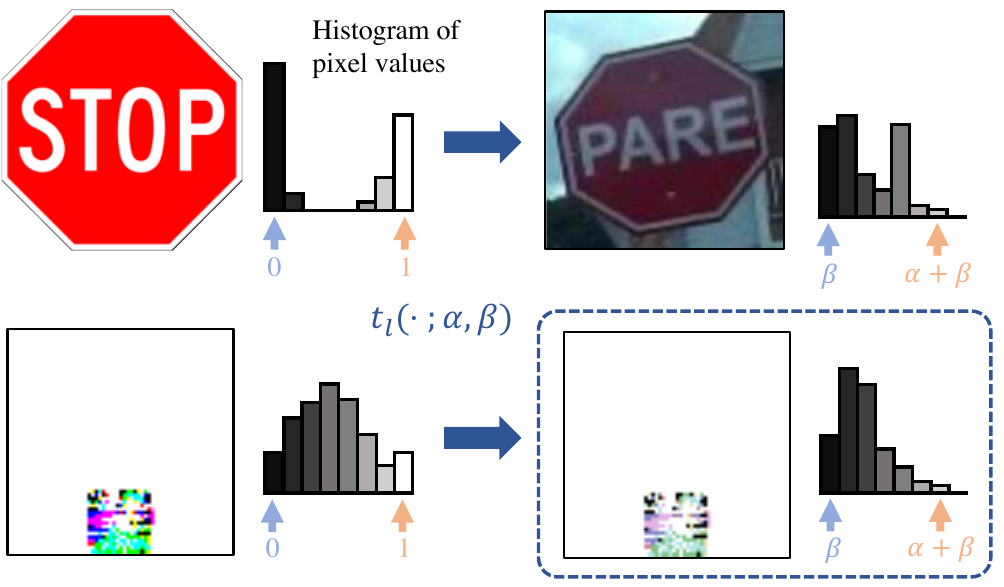}
    \vspace{-20pt}
    \caption{Computing relighting parameters (top) and applying the transform (bottom).}\label{fig:compute_light_tf_steps}
\end{figure}

Each traffic sign in our dataset has two associated relighting parameters, $\alpha, \beta \in \R$.
Given a patch $\tP$, its relighted version $\tP_\text{relighted} = \alpha \tP + \beta$ is rendered on the scene as depicted on the bottom row of \cref{fig:compute_light_tf_steps}.
We infer $\alpha, \beta$ by matching the histogram of the original sign (e.g., the real stop sign on the upper-right of \cref{fig:compute_light_tf_steps}) to a canonical image (e.g., the synthetic stop sign on the upper-left):
in particular, we set $\beta$ as the $p$-th percentile of all the pixel values (aggregated over all three RGB channels) on that sign and $\alpha$ as the difference between the $p$-th and ($1-p$)-th percentile.
We call this the ``percentile'' method and explain why we chose it in \cref{ssec:realism_test}.
This method assumes that relighting can be approximated with a linear transform where $\alpha$ and $\beta$ represent contrast and brightness adjustments.
Like before, since this transformation is linear, it is differentiable.

\subsection{Realism Test}\label{ssec:realism_test}

\begin{table}[t]
    \caption{Comparison of different geometric and relighting transforms from our realism test (mean $\pm$ standard deviation of RMSE across 44 samples). The best results are in bold.}\label{tab:realism_main}
    \vspace{-5pt}
    \small
    \centering
    \begin{tabular}{@{}lllr@{}}
        \toprule
        Transforms                  & Methods                         & Colors & RMSE ($\downarrow$)        \\ \midrule
        \multirow{3}{*}{Geometric}  & Translate \& Scale              & n/a    & 1.72 $\pm$ 1.19            \\
                                    & Affine                          & n/a    & 1.35 $\pm$ 0.49            \\
                                    & Perspective (3D)                & n/a    & \textbf{1.13 $\pm$ 0.41}   \\ \cmidrule(l){2-4}
        \multirow{8}{*}{Relighting} & \multirow{3}{*}{Percentile}     & RGB    & \textbf{0.110 $\pm$ 0.034} \\
                                    &                                 & HSV    & 0.227 $\pm$ 0.118          \\
                                    &                                 & LAB    & 0.652 $\pm$ 0.112          \\ \cmidrule(l){3-4}
                                    & \multirow{3}{*}{Polynomial}     & RGB    & 0.113 $\pm$ 0.037          \\
                                    &                                 & HSV    & 0.118 $\pm$ 0.035          \\
                                    &                                 & LAB    & 0.161 $\pm$ 0.043          \\ \cmidrule(l){3-4}
                                    & \multirow{2}{*}{Color Transfer} & HSV    & 0.117 $\pm$ 0.035          \\
                                    &                                 & LAB    & 0.184 $\pm$ 0.062          \\ \bottomrule
    \end{tabular}
    \vspace{-5pt}
\end{table}

\begin{figure}[t]
    \centering
    \includegraphics[width=0.8\linewidth]{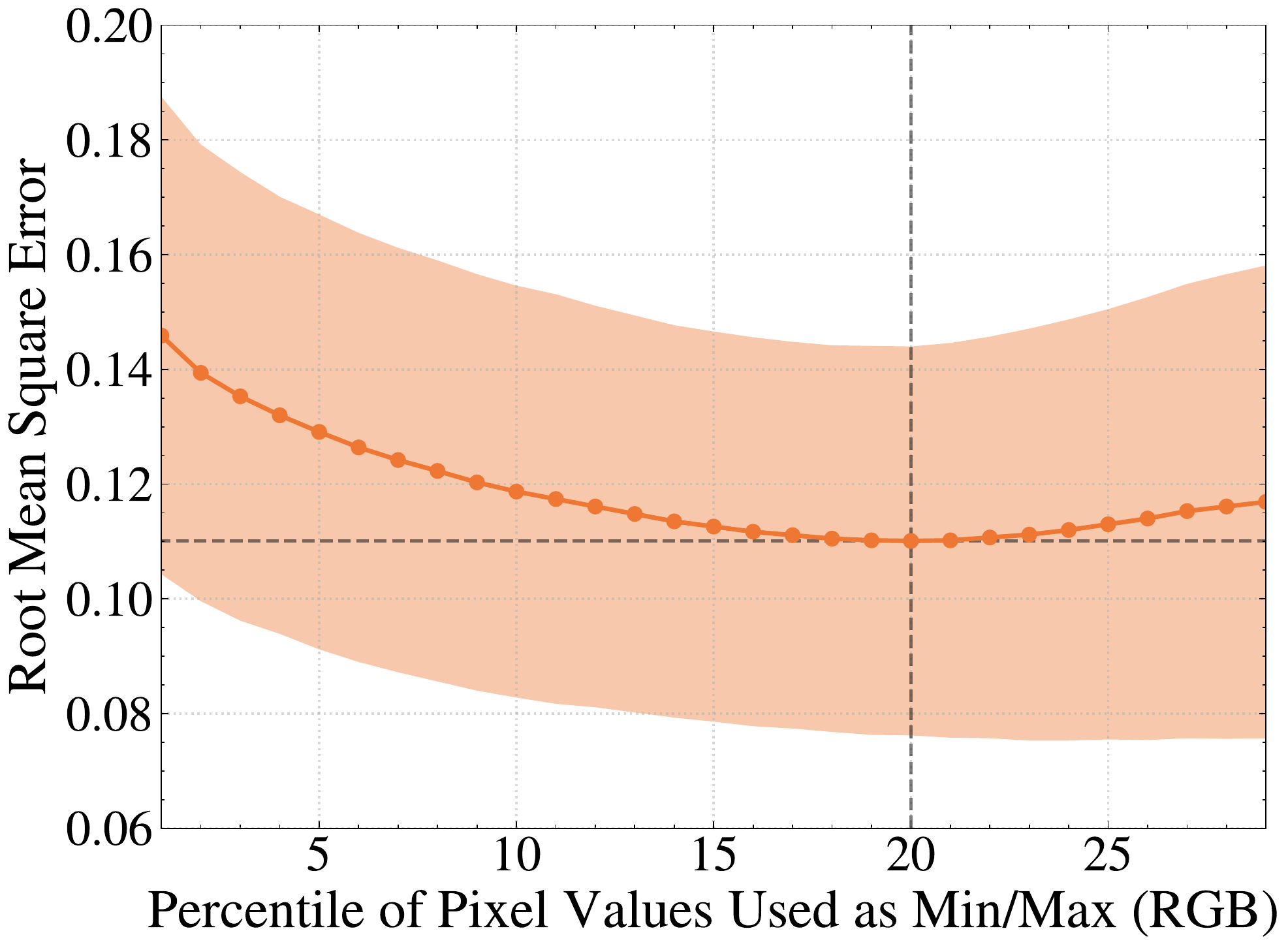}
    \caption{RMSE between the photographed and the rendered patches using the ``percentile'' method with different values of $p$. The shaded region denotes the standard deviation across 44 samples. $p=20$ yields the lowest RMSE.}\label{fig:realism_test_percentile_rgb}
    \vspace{-5pt}
\end{figure}

\begin{figure}[t]
    \centering
    \includegraphics[width=\linewidth]{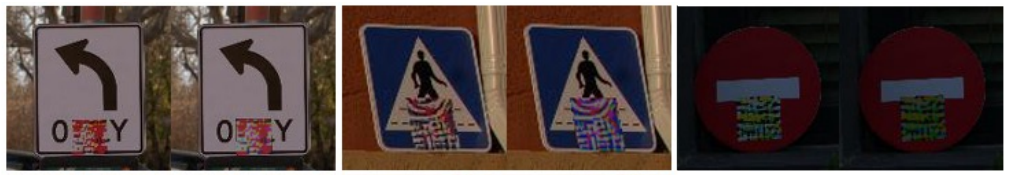}
    \vspace{-20pt}
    \caption{Random samples used in our realism experiments (left: real, right: rendered). \cref{fig:realism_test_all} contains all samples.}\label{fig:real_vs_render}
    \vspace{-10pt}
\end{figure}

In this section, we measure how realistic the patches are when rendered with different transform methods: three for geometric and eight for relighting.
The geometric transforms include perspective (or homographic), affine, and translate \& scale transforms.
For relighting, we experiment with three methods, each of which can be carried out on different color spaces (RGB, HSV, and LAB):
the percentile method, described in \cref{sssec:light};
polynomial fitting, where we find the polynomial that best fits the pixel values on each real sign given the corresponding pixel values on the digital one;
and \emph{Color Transfer}~\citep{reinhard_color_2001}, which tries to match the mean and the standard deviation of the pixel values.

We photograph 44 pairs of \emph{real} traffic signs with and without an adversarial patch: 11 signs, one for each class, in four scenes and lighting conditions.
For each sample, we hand-annotated the keypoints of both the signs and the patches.
Then, given an image of the sign without the patch, we render the patch on it using the different transform methods.
For geometric transforms, we measure the root mean square error (RMSE) between the rendered and the corresponding groundtruth \emph{corners} of the patch.
To compare the relighting methods, we compute the RMSE between the rendered and the corresponding groundtruth \emph{pixel values} of the patch.
We use the groundtruth geometric transform when computing the relighting parameters to disentangle the potential error from the geometric transform.

\cref{tab:realism_main} reports the best RMSE achieved by each transform after a hyperparameter sweep ($p$, polynomial degrees, etc.).
The perspective transform achieves the lowest RMSE as expected which emphasizes the importance of using the full 3D transform instead of simpler alternatives.
For relighting, the percentile method with $p=0.2$ performs as well as, or better than, any other at rendering the adversarial patches.
Hence, these are the two transforms we use to construct the \apb{} benchmark and in all of the experiments in \cref{sec:result} unless stated otherwise.
\cref{fig:real_vs_render} visually compares the rendered patches with the groundtruth ones under these best transform methods.
\cref{ap:sec:realism_test} contains more details.

\section{Experiments on \apb{} Benchmark}\label{sec:result}

Our benchmark can be used to evaluate attacks and defenses under various threat models, e.g., making objects appear vs disappear, using a universal patch vs a targeted attack, etc.
In this paper, we focus on the setting where the adversary tries to make a traffic sign \emph{disappear} or be \emph{misclassified} using the \emph{per-class} attack, i.e., only one patch per class of objects, similar to \citet{benz_universal_2021}.
We argue that this threat model is more realistic and more alarming as the attacker only needs to distribute several adversarial stickers that are effective across million of traffic signs.
We assume the adversary has access to the target model (white-box).

\subsection{Experiment Setup}\label{ssec:setup}

\paragraph{Traffic sign detectors.}
We experiment with three object detection models, Faster R-CNN~\citep{ren_faster_2015}, YOLOF~\citep{chen_you_2021}, and DINO~\citep{zhang_dino_2022}, all trained on the MTSD dataset to predict bounding boxes for all 11 traffic sign classes plus the ``other'' class.
We follow the training method and hyperparameters from \citet{neuhold_mapillary_2017}.
As mentioned in \cref{ssec:metrics}, we report the false negative rate (FNR) in addition to mAP scores.
For FNR, the score threshold is chosen as one that maximizes the F1 score on the validation set of MTSD.

\paragraph{Attack algorithms.}
We use the RP2 attack~\citep{eykholt_physical_2018} and the DPatch attack~\citep{liu_dpatch_2019} to generate adversarial patches for all models.
We assume that the adversary has access to 5 held-out images from our benchmark and use them to generate \textbf{one adversarial patch per sign class}.
We note that this setting, referred to as \textit{per-class} attack, is different from the usual white-box attack threat model where each sample is given a unique perturbation (we call it \textit{per-instance} attack) and is more similar to the ``universal'' adversarial perturbation~\citep{moosavi-dezfooli_universal_2017}.
We argue that this threat model is more realistic and more alarming as the attacker only needs to distribute one adversarial sticker that are effective across million of traffic signs of the same type.
\cref{ap:ssec:per_instance} discusses the threat models in more detail.

Each of the classes has a specific set of these 5 images each of which contains at least one sign of that class.
For \apbs{}, we use 50 images since there are more samples per class.
In practice, an adversary may benefit from using more than 5 (or 50) images to generate the patch, but we set our limit here to leave sufficient samples for the evaluation phase.
We do not find a significant difference in the performance of the two attacks (\cref{ap:ssec:rp2_vs_dpatch}) so we report only the results of DPatch attack with PGD optimizer in the main paper.

\paragraph{Defense algorithm.}
We use adversarial training with DPatch attack and five-step PGD as it performs the best empirically.
The patches are generated \emph{per-instance} at a random location to prevent overfitting to a specific one.
To improve the effectiveness of adversarial training under a small number of steps, we cache patches generated in the previous epoch and use it as an initialization for the next one~\citep{zheng_efficient_2020}.
For more detailed setup and results, please see \cref{ap:sec:exp_setup}.

\paragraph{Synthetic Benchmark.}
We use canonical synthetic signs, one per class, as a baseline to compare our \apbs{} benchmark to (we cannot find canonical synthetic signs for all 100 classes of \apb{}).
Similarly to \citet{eykholt_physical_2018}, the synthetic sign is placed at a random location on one of 50 random background images and randomly rotated between 0 and 15 degrees.
We use the synthetic benchmark for both generating and testing the adversarial patch.
For testing, we use 2,000 background images per class, randomly selected from our \apbs{} benchmark to keep the distribution of the scenes similar.

\subsection{Evaluation Metrics}\label{ssec:metrics}

Here, we define \emph{a successful attack} as a patch that makes the sign either (i) undetected or (ii) classified to a wrong class (i.e., any of the other classes, or the background class).
Similarly to previous work, we measure the effectiveness of an attack by the \emph{attack success rate} (ASR), defined as follows.
Given a list of signs $\{x_i\}_{i=1}^N$ and the corresponding version with an adversarial patch applied to it, $\{x'_i\}_{i=1}^N$,
\begin{align}
    \mathrm{ASR} = \frac{\sum_{i=1}^N \1_{x_i \mathrm{ is~detected}} ~\land~ \1_{x'_i \mathrm{ is~not~detected}}}{\sum_{i=1}^N \1_{x_i \mathrm{is~detected}}}.
\end{align}
ASR and FNR are easy to interpret but dependent on specific thresholds of both the confidence score and the IoU between the groundtruth and the detected boxes.
Hence, we also report mAP which averages across these thresholds.

\subsection{Main Results}\label{ssec:results}

\begin{table*}[t]
    \centering
    \caption{Mean FNR and mAP of the adversarial patches on six traffic sign detectors on \apb{}. ``FRCNN'' refers to Faster R-CNN, ``Adv.'' indicates adversarially trained models. For defenders, lower FNR ($\downarrow$) and higher mAP ($\uparrow$) are better.
    }
    \vspace{-5pt}\label{tab:reap_avg_main}
    \small
    \begin{tabular}{@{}lrrrrrrrrrrrr@{}}
        \toprule
        \multirowcell{2}[0pt][l]{Patch Size}   & \multicolumn{2}{c}{FRCNN} & \multicolumn{2}{c}{YOLOF} & \multicolumn{2}{c}{DINO} & \multicolumn{2}{c}{Adv. FRCNN} & \multicolumn{2}{c}{Adv. YOLOF} & \multicolumn{2}{c}{Adv. DINO}
        \\ \cmidrule(lr){2-3} \cmidrule(lr){4-5} \cmidrule(lr){6-7} \cmidrule(lr){8-9} \cmidrule(lr){10-11} \cmidrule(lr){12-13}
                                               & FNR                       & mAP                       & FNR                      & mAP                            & FNR                            & mAP                           & FNR  & mAP  & FNR  & mAP  & FNR & mAP  \\ \midrule
        No patch                               & 4.3                       & 72.9                      & 18.5                     & 54.8                           & 14.1                           & 68.2                          & 3.1  & 73.3 & 21.0 & 55.0 & 9.4 & 74.2 \\
        Small (\psize{10}{10})                 & 15.4                      & 59.4                      & 33.7                     & 43.5                           & 32.0                           & 60.4                          & 3.8  & 71.8 & 22.5 & 54.7 & 1.8 & 80.6 \\
        Medium (\inch{10}$\times$\inch{20})    & 22.4                      & 46.5                      & 42.7                     & 36.6                           & 35.4                           & 52.6                          & 6.1  & 66.8 & 27.1 & 51.9 & 1.2 & 80.1 \\
        Large (two \inch{10}$\times$\inch{20}) & 50.0                      & 18.2                      & 72.8                     & 19.4                           & 62.8                           & 39.5                          & 13.9 & 56.3 & 57.7 & 34.1 & 3.6 & 77.8 \\
        \bottomrule
    \end{tabular}
\end{table*}

\begin{table*}[t]
    \centering
    \caption{Mean ASR and FNR of the adversarial patches on the six traffic sign detectors on \apbs{}. For sign-specific metrics, see \cref{tab:10x10_syn_vs_real_all_classes}. It is clear that evaluating on the synthetic signs overestimates the attack's potency in every setting. For defenders, lower FNR ($\downarrow$) and ASR ($\downarrow$) are better.
    }
    \vspace{-5pt}\label{tab:syn_vs_real_avg_main}
    \small
    \begin{tabular}{@{}llrrrrrrrrrrrr@{}}
        \toprule
        \multirowcell{2}[0pt][l]{Patch Size} & \multirowcell{2}[0pt][l]{Benchmarks} & \multicolumn{2}{c}{FRCNN} & \multicolumn{2}{c}{YOLOF} & \multicolumn{2}{c}{DINO} & \multicolumn{2}{c}{Adv. FRCNN} & \multicolumn{2}{c}{Adv. YOLOF} & \multicolumn{2}{c}{Adv. DINO}
        \\ \cmidrule(lr){3-4} \cmidrule(lr){5-6} \cmidrule(lr){7-8} \cmidrule(lr){9-10} \cmidrule(lr){11-12} \cmidrule(lr){13-14}
                                             &                                      & FNR                       & ASR                       & FNR                      & ASR                            & FNR                            & ASR                           & FNR  & ASR  & FNR  & ASR  & FNR & ASR \\ \midrule
        \multirowcell{2}[0pt][l]{No patch}   & Synthetic                            & 19.8                      & n/a                       & 17.0                     & n/a                            & 12.7                           & n/a                           & 15.7 & n/a  & 19.0 & n/a  & 5.8 & n/a \\
                                             & \apbs{} (ours)                       & 20.2                      & n/a                       & 17.1                     & n/a                            & 12.8                           & n/a                           & 17.4 & n/a  & 19.2 & n/a  & 6.1 & n/a \\
        \cmidrule(){2-14}
        \multirowcell{2}[0pt][l]{Small                                                                                                                                                                                                                                                                           \\(\psize{10}{10})} & Synthetic & 76.9 & 73.1 & 89.8 & 88.6 & 58.8 & 56.9 & 50.0 & 43.9 & 76.8 & 73.4 & 24.1 & 22.6        \\
                                             & \apbs{} (ours)                       & 50.5                      & 39.2                      & 48.6                     & 38.9                           & 36.2                           & 28.0                          & 18.7 & 5.1  & 28.3 & 12.7 & 1.1 & 0.1 \\
        \cmidrule(){2-14}
        \multirowcell{2}[0pt][l]{Medium                                                                                                                                                                                                                                                                          \\(\inch{10}$\times$\inch{20})} & Synthetic &  89.9 & 88.3 & 92.0 & 91.1 & 73.1 & 72.6 & 79.5 & 77.3 & 83.7 & 81.7 & 34.7 & 33.9       \\
                                             & \apbs{} (ours)                       & 64.4                      & 56.1                      & 60.5                     & 52.8                           & 45.5                           & 38.4                          & 33.8 & 23.5 & 46.5 & 34.7 & 1.3 & 0.1 \\
        \cmidrule(){2-14}
        \multirowcell{2}[0pt][l]{Large                                                                                                                                                                                                                                                                           \\(two \inch{10}$\times$\inch{20})} & Synthetic & 99.6	 & 99.6 & 100.0 & 100.0 & 96.5 & 96.4 & 98.9 & 98.8 & 99.1 & 98.9 & 52.9 &  52.7       \\
                                             & \apbs{} (ours)                       & 85.2                      & 82.0                      & 88.2                     & 86.1                           & 85.1                           & 83.5                          & 59.3 & 53.6 & 69.9 & 64.3 & 5.1 & 4.3 \\
        \bottomrule
    \end{tabular}
\end{table*}

\begin{figure*}[t]
    \centering
    \includegraphics[width=1.0\textwidth]{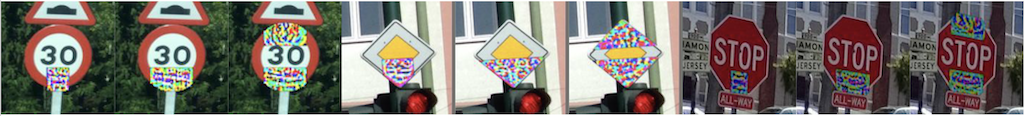}
    \vspace{-15pt}
    \caption{Examples of small (\psize{10}{10}), medium (\psize{10}{20}) and large (two \psize{10}{20}) patches applied to three of the signs from our benchmark. The large patch size is clearly visible and obscures the notion of imperceptibility. We still choose to experiment with it since it is approximately the same size used by prior work~\citep{eykholt_physical_2018,zhao_seeing_2019}.}\label{fig:patch_size_example}
\end{figure*}

Experiments on our \apb{} benchmark illuminate several findings that were not previously observed due to the lack of scalability and reproducibility of real-world experiments:

\paragraph{(1) Patch attacks against road signs are less effective than previously believed.}
From \cref{tab:reap_avg_main}, a \psize{10}{10} adversarial patch increases FNR by only 11--18 percentage points on the undefended models.
For \apbs{} (\cref{tab:syn_vs_real_avg_main}), the increase is 8--12 percentage points.
For comparison, a class-wise adversarial perturbation under imperceptible $\ell_\infty$ norms achieves above 90\% success rate~\citep{benz_universal_2021}.
More importantly, \textbf{on adversarially trained models, FNR remains almost identical before and after applying the patch} on Faster R-CNN and YOLOF.
Surprisingly, adversarially trained DINO performs better on samples with adversarial patches than without.
We hypothesize that this is a result of overfitting to some adversarial patches and not a clear sign of weak attacks or gradient obfuscation.
We refer to \cref{ssec:more_atk} for additional detail.

This result implies that a well-known defense like adversarial training is effective and may be sufficient to protect against patch attacks in the real world.
Adversarial attacks are most troubling when they are imperceptible; patches as large as \psize{10}{10} (or larger) are likely to draw attention, which may make them less of a threat in practice.

Our findings are also consistent with prior works that investigate physical-world attacks on stop signs.
In these works, the attack is often clearly visible.
For instance, \citet{eykholt_physical_2018} and \citet{zhao_seeing_2019} use a patch that is close to our two \inch{10}$\times$\inch{20} patches which is why they observe a high attack success rate similar to our results with the larger patch size.
Nonetheless, a patch of this size surely breaches all notions of imperceptibility.
Perhaps an interesting threat model to study in the future is to allow large patches but additionally constrain the perturbation with $\ell_\infty$-norm.

\begin{figure}[t]
    \centering
    \begin{subfigure}[b]{0.22\textwidth}
        \centering
        \includegraphics[width=\textwidth]{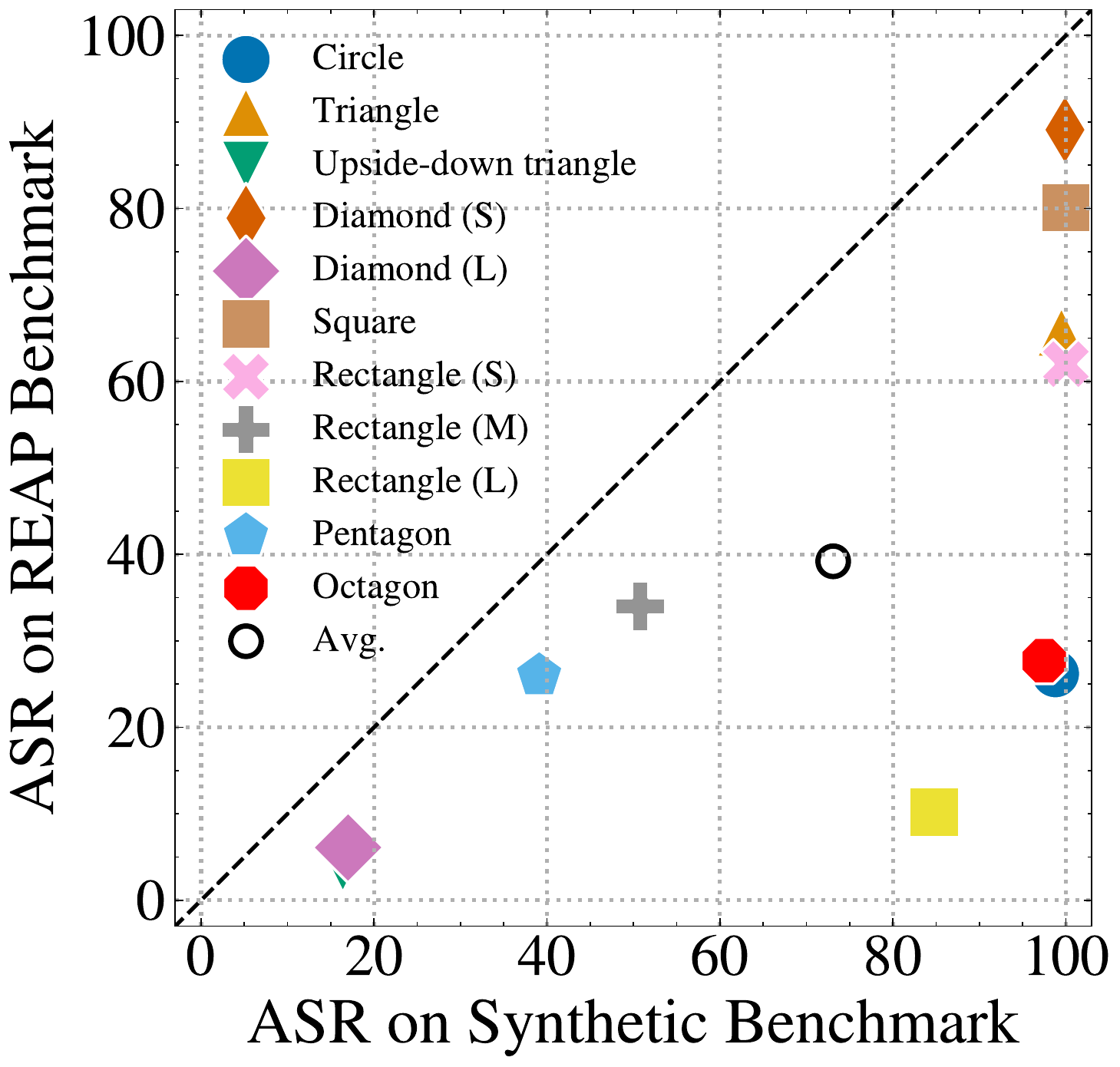}
        \caption{Faster R-CNN}\label{fig:reap_vs_syn_frcnn}
    \end{subfigure}
    \hfill
    \begin{subfigure}[b]{0.22\textwidth}
        \centering
        \includegraphics[width=\textwidth]{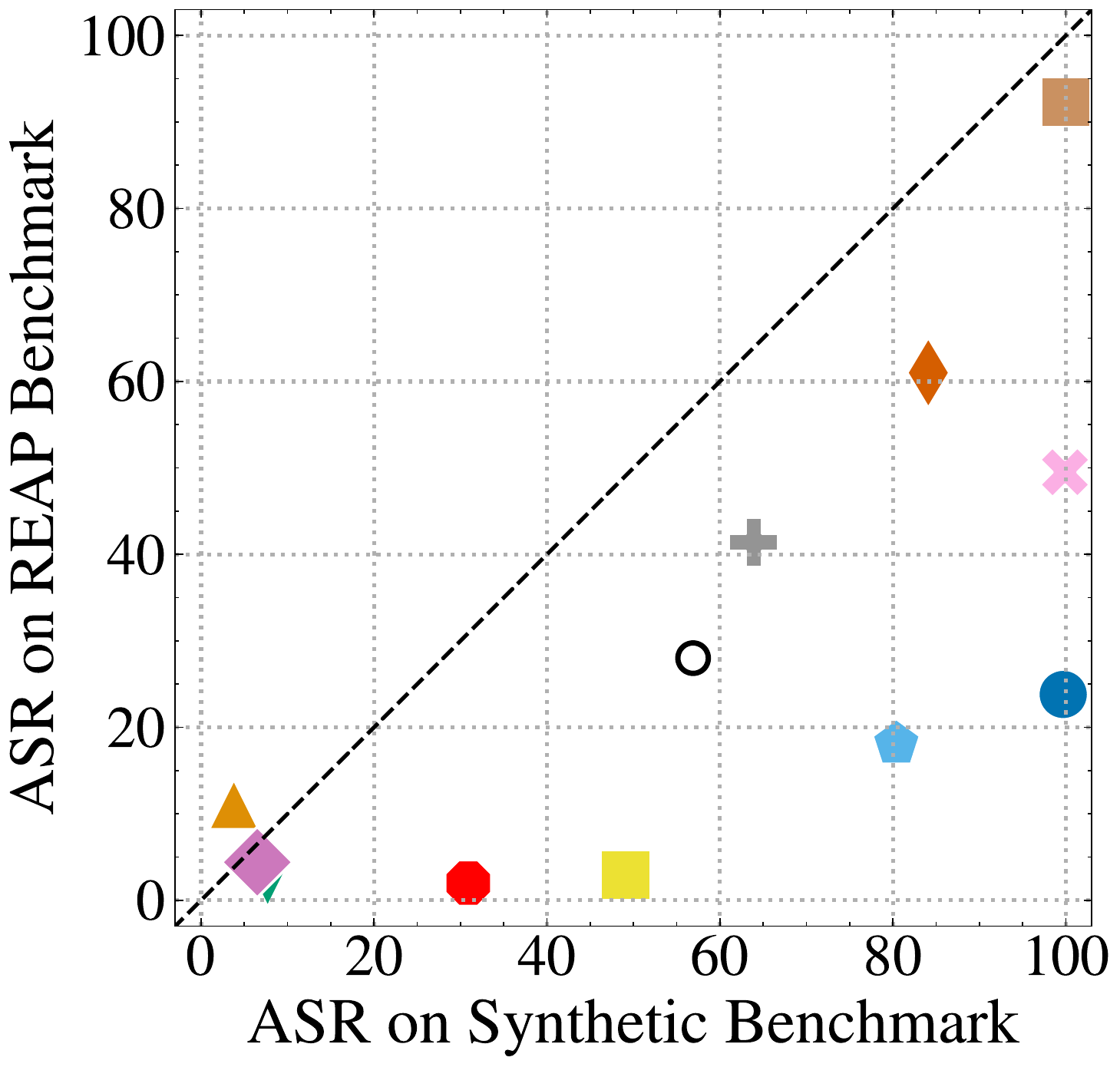}
        \caption{DINO}\label{fig:reap_vs_syn_dino}
    \end{subfigure}
    \vspace{-5pt}
    \caption{ASRs on synthetic vs \apbs{} benchmarks for Faster R-CNN (left) and DINO (right). The dashed line marks the points with an equal ASR on both datasets.}\label{fig:reap_vs_syn}
    \vspace{-5pt}
\end{figure}

\paragraph{(2) ASR measured on synthetic data is not predictive of ASR measured on our realistic benchmark.}
We compare \apbs{} to a synthetic benchmark intended to be representative of methodology often found in prior work: we take a single synthetic image of a road sign, then generate attacks against it (instead of a real image).
\cref{tab:syn_vs_real_avg_main} and \cref{fig:reap_vs_syn} show that there is a large difference between metrics as measured on such a synthetic benchmark compared to our benchmark.
The gap can be up to 50--60 percentage points on average.

\cref{fig:reap_vs_syn} and \cref{tab:10x10_syn_vs_real_all_classes} in \cref{ap:sec:more_exp} compare ASR on the two benchmarks by class of the traffic signs.
If the two ASRs were similar, all data points would have lied close to the diagonal dashed line.
Instead, most of the data points are below the line, suggesting that the synthetic benchmark consistently overestimates the ASR.
Moreover, there is no clear relationship between the two measurements of ASR.
If the \emph{rankings} of the ASRs are well-correlated, we should expect ordering of the points to be similar in both horizontal and vertical directions, but this is not the case.

\begin{table}[t]
    \centering
    \caption{Robustness of the adversarially trained models under different attack threat models (\psize{10}{10} patch size). The per-instance attack has the highest ASR and the lowest mAP as expected, and there is no sign of gradient obfuscation.}\label{tab:transfer}
    \vspace{-5pt}
    \small
    \begin{tabular}{@{}lrr@{}}
        \toprule
        Attacks                         & ASR ($\uparrow$) & mAP ($\downarrow$) \\ \midrule
        \multicolumn{3}{c}{Adv. Faster R-CNN}                                   \\
        \midrule
        No Attack                       & n/a              & 66.0               \\
        Per-Class Attack                & 5.1              & 65.7               \\
        Per-Instance Attack             & \textbf{16.0}    & \textbf{59.3}      \\
        Transfer from YOLOF             & 3.2              & 67.8               \\
        Transfer from Adv. YOLOF        & 7.0              & 63.6               \\
        Transfer from Synthetic         & 2.7              & 69.1               \\
        \midrule
        \multicolumn{3}{c}{Adv. YOLOF}                                          \\
        \midrule
        No Attack                       & n/a              & 58.5               \\
        Per-Class Attack                & 17.7             & 51.3               \\
        Per-Instance Attack             & \textbf{28.2}    & \textbf{46.5}      \\
        Transfer from Faster R-CNN      & 13.5             & 53.1               \\
        Transfer from Adv. Faster R-CNN & 7.9              & 55.4               \\
        Transfer from Synthetic         & 12.2             & 54.3               \\
        \midrule
        \multicolumn{3}{c}{Adv. DINO}                                           \\
        \midrule
        No Attack                       & n/a              & 65.7               \\
        Per-Class Attack                & 0.1              & 75.1               \\
        Per-Instance Attack             & \textbf{2.7}     & \textbf{63.7}      \\
        Transfer from Adv. Faster R-CNN & 0.1              & 76.5               \\
        Transfer from Adv. YOLOF        & 0.2              & 76.1               \\
        Transfer from DINO              & 0.0              & 79.6               \\
        Transfer from Synthetic         & 0.4              & 72.7               \\
        \bottomrule
    \end{tabular}
\end{table}

\begin{figure}[t]
    \centering
    \includegraphics[width=0.95\linewidth]{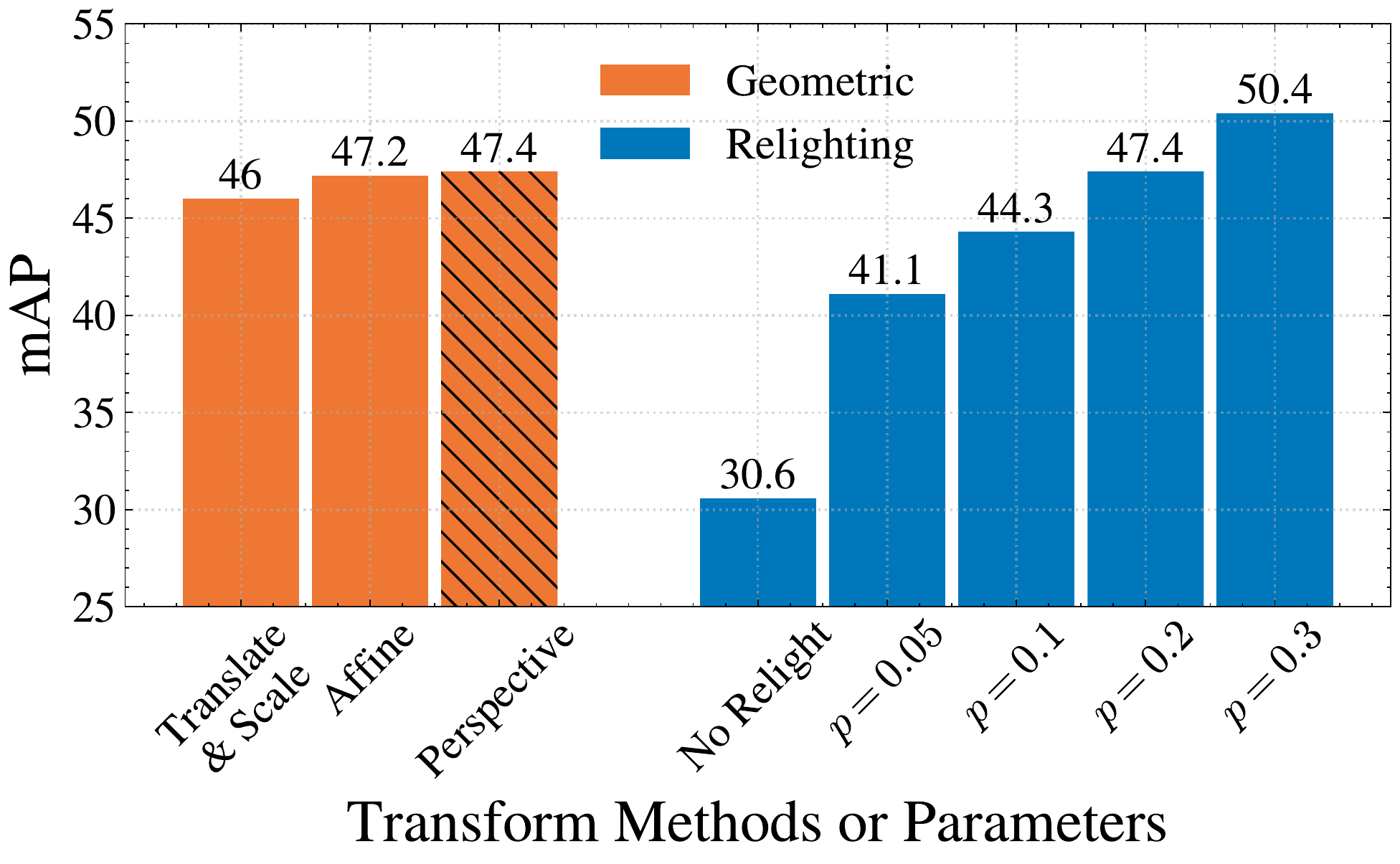}
    \vspace{-5pt}
    \caption{Effects of different geometric and relighting transform methods on the mAP of the YOLOF model under attack. The hatched bars are the default setting. When either geometric or relighting transform is varied, the other one is fixed to the default (perspective, $p=0.2$).}\label{fig:transforms_mAP}
    \vspace{-5pt}
\end{figure}

\paragraph{(3) The lighting transform affects the attack's effectiveness more than the geometric transform.}
\cref{fig:transforms_mAP} shows how the transformations our benchmark applies to the patch affect its mAP scores.
For all models, our realistic lighting transform has a much larger effect than the geometric transform.
Without the lighting transform, mAP decreases by 17 percentage points for YOLOF and 15 for Faster R-CNN (i.e., an increase of 23 and 14 points on the ASR).
This observation explains why the synthetic benchmark as well as synthetic evaluations in previous works overestimate ASR.

\subsection{Extended Attack Evaluation}\label{ssec:more_atk}
\begin{table}[t]
    \centering
    \small
    \setlength\tabcolsep{3.5pt}
    \begin{tabular}{@{}llrrrrrr@{}}
        \toprule
        \multirowcell{2}[0pt][l]{Augment}      & \multirowcell{2}[0pt][l]{Strength} & \multicolumn{2}{c}{FRCNN} & \multicolumn{2}{c}{Adv. YOLOF} & \multicolumn{2}{c}{Adv. DINO}
        \\ \cmidrule(lr){3-4} \cmidrule(lr){5-6} \cmidrule(lr){7-8}
                                               &                                    & FNR                       & mAP                            & FNR                           & mAP  & FNR & mAP  \\ \midrule
        None                                   & n/a                                & 15.4                      & 59.4                           & 22.5                          & 54.7 & 1.2 & 80.1 \\ \cmidrule{2-8}
        \multirowcell{3}[0pt][l]{Color-jitter} & 0.1                                & 16.1                      & 58.8                           & 23.1                          & 54.7 & 1.2 & 80.1 \\
                                               & 0.2                                & 16.0                      & 58.5                           & 23.6                          & 54.6 & 1.2 & 80.0 \\
                                               & 0.3                                & 15.5                      & 59.0                           & 23.3                          & 54.7 & 1.3 & 80.1 \\ \cmidrule{2-8}
        \multirowcell{3}[0pt][l]{Unif. noise}  & 0.1                                & 15.7                      & 58.9                           & 23.0                          & 54.8 & 1.2 & 80.4 \\
                                               & 0.2                                & 15.4                      & 58.8                           & 22.6                          & 54.7 & 1.4 & 80.1 \\
                                               & 0.3                                & 15.6                      & 58.6                           & 22.9                          & 54.7 & 1.4 & 80.2 \\
        \bottomrule
    \end{tabular}
    \caption{FNR and mAP with color-jitter or random noise applied during the attack EoT on \apb{}. We use the small patch for Faster R-CNN and Adv. YOLOF, and the medium size for Adv. DINO. None of the augmentations seems to affect the potency of the attack.}\label{tab:aug}
\end{table}

Because these results were so surprising, we investigated the possibility that our attack algorithms are not sufficiently strong (e.g., gradient obfuscation~\citep{athalye_obfuscated_2018}) or that the adversarially trained models ``catastrophically overfit''~\citep{wong_fast_2020,kim_understanding_2021,andriushchenko_understanding_2020}, i.e., they memorize the attack patterns during training but are not actually robust.
In particular, we evaluate the adversarially trained models against \emph{transfer} and \emph{per-instance} attacks.

The per-instance attack generates one patch for each instance of traffic signs, as opposed to our default per-class patch.
The transfer attack generates per-class patches from either a different source model or the synthetic data.
\cref{tab:transfer} shows that the per-instance attack always achieves a higher ASR (and lower mAP) than the per-class attack, and the transfer attack has the lowest ASR in most cases.
This result is expected and does not indicate that the gradient obfuscation or the catastrophic overfitting phenomenon is happening.

To further improve the robustness of the adversarial patches (i.e., making the attack transfer to other instances in the same class), we also tried to generate the adversarial patches by applying random augmentations including color-jitter and uniform noise injection (similar to expectation over transformation~\citep{athalye_synthesizing_2018}).
\cref{tab:aug} shows that the augmentations with varying strength levels do not significantly affect the ASR of the attack.

Overall, the effectiveness of all the attacks remains limited against all the adversarially trained models.
In particular, the ASR of the per-instance attack, which is an \emph{upper bound} of ASR on all threat models, is only 3\% on Adv. DINO on \apbs{}.
Based on these experiments (and others in \cref{ap:ssec:cover_all}), we tentatively conclude that the adversarially trained classifiers truly do appear robust for the REAP detection task.
Because this result is so surprising, further research is needed before we can have full confidence in this conclusion.

\section{Conclusion and Future Directions}

We construct the first large-scale benchmark for evaluating adversarial patches.
Our benchmark consists of over 14,000 signs from real driving scenes, and each sign is annotated with the transformations necessary to render an adversarial patch realistically onto it.
Using this benchmark, we experiment with a broad range of models and attacks.
We find that adversarial patches of a clearly visible size fool an undefended model on less than 28\% of the signs and only 1\% for a defended model.
This is in contrast to adversarial examples with bounded $\ell_p$-norm, where attacks nearly always succeed.
\textbf{All in all, our experiments suggest that realistic constraints render patch attacks significantly less effective, and vanilla adversarial training is an effective defense against the current practical patch attacks.}

One interesting direction for future research is to explore whether attacks against object detectors can be improved.
Also, in our experiments, adversarial training achieved strong robustness at the cost of degrading mAP on clean images by about 5 percentage points.
It would be interesting to explore new defenses that have less impact on clean performance.
We hope that our benchmark will provide a foundation for more realistic evaluation of patch attacks and drive future research on defenses against them.

{\small
      \bibliography{reference.bib,reference_extra.bib}
      \bibliographystyle{ieee_fullname}
}

\newpage
\appendix
\onecolumn

\section{Additional Details of \apb{} Benchmark}\label{ap:sec:anno_detail}

\subsection{Basic Usage}\label{ap:ssec:usage}

\apb{} benchmark provides simple tooling for rendering adversarial patch onto desired traffic signs.
We integrate it to \verb|detectron2|\footnote{\url{https://github.com/facebookresearch/detectron2}}~\citep{wu_detectron2_2019}, a popular object detection/segmentation framework, since many repositories and recent research are built on top of \verb|detectron2|.
That said, \apb{} can be used with any other framework including pure PyTorch.
This tool consists of three main components:
\begin{enumerate}[leftmargin=*]
   \item \verb|reap_annotations.csv|: The annotation for each sign in our \apb{} benchmark corresponds to one row in this \verb|csv| file. We load it in as a \verb|pandas.DataFrame| object, and the transform parameters are read and simply appended to the other metadata and labels in each sample (e.g., image height/width, bounding boxes, etc.).
   \item \verb|RenderObject| class: We create one \verb|RenderObject| for each sign we want to apply an adversarial patch to. \verb|RenderObject| holds the parameters of the geometric and the relighting transforms for that sign and also applies the transformations when called through \verb|RenderImage.apply_objects()|. We use two separate subclasses to implement \verb|RenderObject| for the real signs in \apb{} benchmark and for the synthetic data.
   \item \verb|RenderImage| class: We wrap each sample in a \verb|RenderImage| object. \verb|RenderImage| holds the original image and a dictionary of multiple \verb|RenderObject|'s. Once \verb|RenderObject.apply_objects()| is called, it loops through all of its \verb|RenderObject|'s and applies the transformed adversarial patches to the image.
\end{enumerate}

Below we show a snippet of how this tool is generally used for applying patches for the \apb{} benchmark.
This process can be applied during both the attack generation and the evaluation.
Our Github repository (\mylink{https://github.com/wagner-group/reap-benchmark}) contains the code for both evaluating models on \apb{} and (adversarially) training new ones on the MTSD dataset.

\begin{python}
   # Given input parameters
   # An input sample (e.g., in detectron2 format) which already comes loaded
   # with REAP transform parameters handled mostly by a DatasetMapper.
   sample: Dict[str, Any]
   adv_patch: torch.Tensor  # Generated adversarial patch
   patch_mask: torch.Tensor  # Binary mask of adversarial patch
   mode: str  # Benchmark mode (either "reap" or "synthetic")
   obj_class: int  # Target object class to attack/evaluate

   # Create RenderImage object around a given sample
   rimg: RenderImage = RenderImage(sample, mode=mode, obj_class=obj_class)

   # Render adv_patch on rimg and post process it into desired format
   img_render, target_render = rimg.apply_objects(adv_patch, patch_mask)
   img_render = rimg.post_process_image(img_render)

   # Perform inference on rendered image
   outputs: Dict[str, Any] = predict(img_render)

   # Compute metrics comparing outputs to target_render
   ...
\end{python}

The operations on the image and the adversarial patch are implemented with \verb|PyTorch| and the \verb|Kornia| package~\citep{riba_kornia_2020}, which also uses \verb|PyTorch| underneath, so the entire process can be executed on a GPU or a CPU and is entirely differentiable.
Our hardware setup (one Nvidia Tesla V100 with six CPU cores) can evaluate anywhere between one to two images per second with the default resolution of 1536$\times$2048 pixels.
This includes data loading, preprocessing, applying a patch to at least one sign on that image, and running an inference.
The total evaluation of our \apb{} benchmark is about five hours or less.
The total time depends mostly on the number of CPU cores (e.g., using 16 vs 8 CPU cores cuts down the total runtime by about half) and less on the GPU specs since we evaluate one image at a time and not in batch.
In the next section (\cref{ap:ssec:apply_patch}), we provide additional details of how the transforms are applied and what is going on inside of \verb|RenderObject|.

\subsection{Applying the Transforms} \label{ap:ssec:apply_patch}

\begin{figure}[t]
   \centering
   \includegraphics[width=0.97\textwidth]{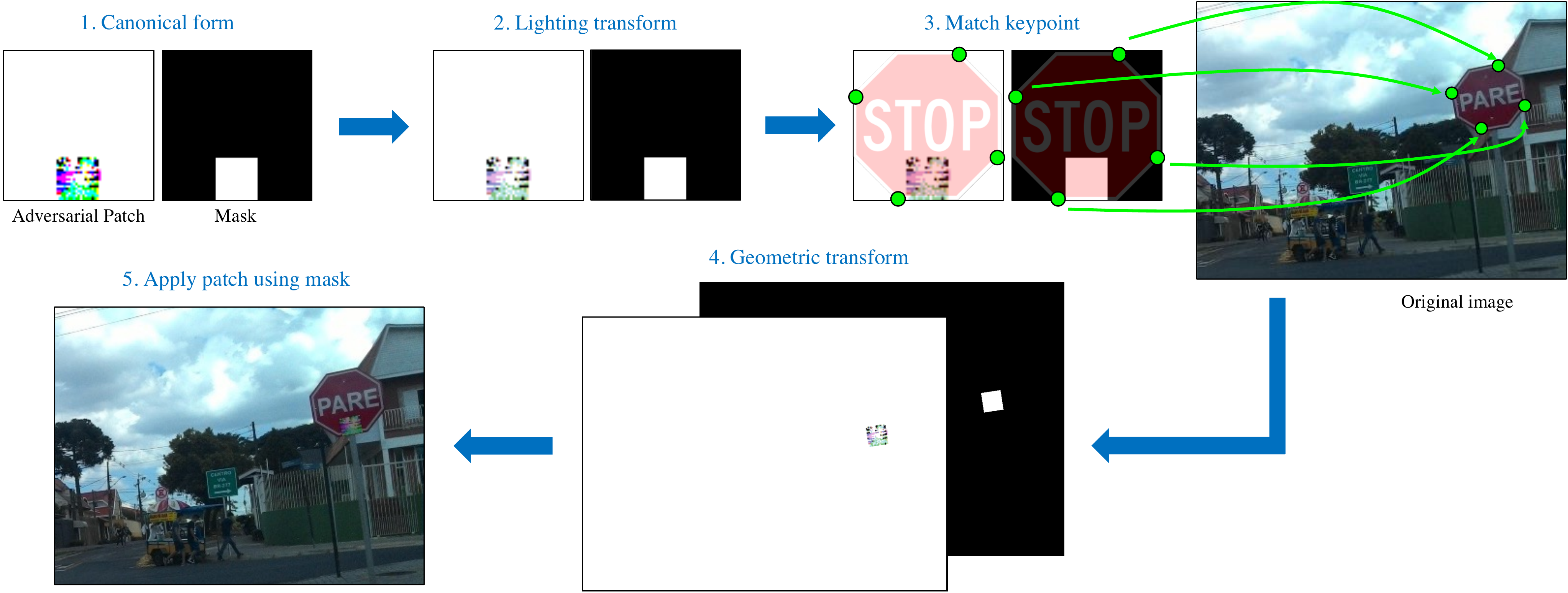}
   \vspace{-5pt}
   \caption{\apb's procedure for applying lighting and geometric transforms to a digitally generated adversarial patch.}\label{fig:diagram}
\end{figure}

\cref{fig:diagram} summarizes the steps to apply an adversarial patch using our \apb{} benchmark.
Given a patch and a corresponding mask with respect to the canonical sign, we first apply relighting transform on the patch.
Then, we use the annotated keypoints of the target sign to determine the parameters of the perspective transform which is then applied to both the patch and the mask.
Throughout this paper, we use bilinear interpolation for any geometric transform.
Finally, the transformed patch is applied to the image using the transformed mask.

To be precise, let $\mX$, $\mP$, and $\mM$ denote the original image, the adversarial patch, and the patch mask, respectively.
The final image $\mX'$ is obtained by the following equation
\begin{align}
   \mX' & = t_g\left(\mM\right) \odot t_g\left( t_l\left(\mP\right)\right) + \left(1 - t_g\left(\mM\right) \right) \odot \mX
\end{align}
where $t_g(\cdot)$ and $t_l(\cdot)$ are the geometric and the relighting transforms which in fact, depend on the annotated parameters associated with $\mX$.

We note that the mask is concatenated to the patch, and both are applied with the same geometric transform and interpolation.
Therefore, $t_g\left(\mM\right)$ is no longer a binary mask like $\mM$.
This creates an effect where the transformed patch blends in more cleanly with the sign than the nearest interpolation does.
Additionally, we also clip the pixel values after applying each transform to ensure that they always stay between 0 and 1.

\begin{figure}[t]
   \centering
   \begin{minipage}{.58\textwidth}
      \centering
      \includegraphics[width=\textwidth]{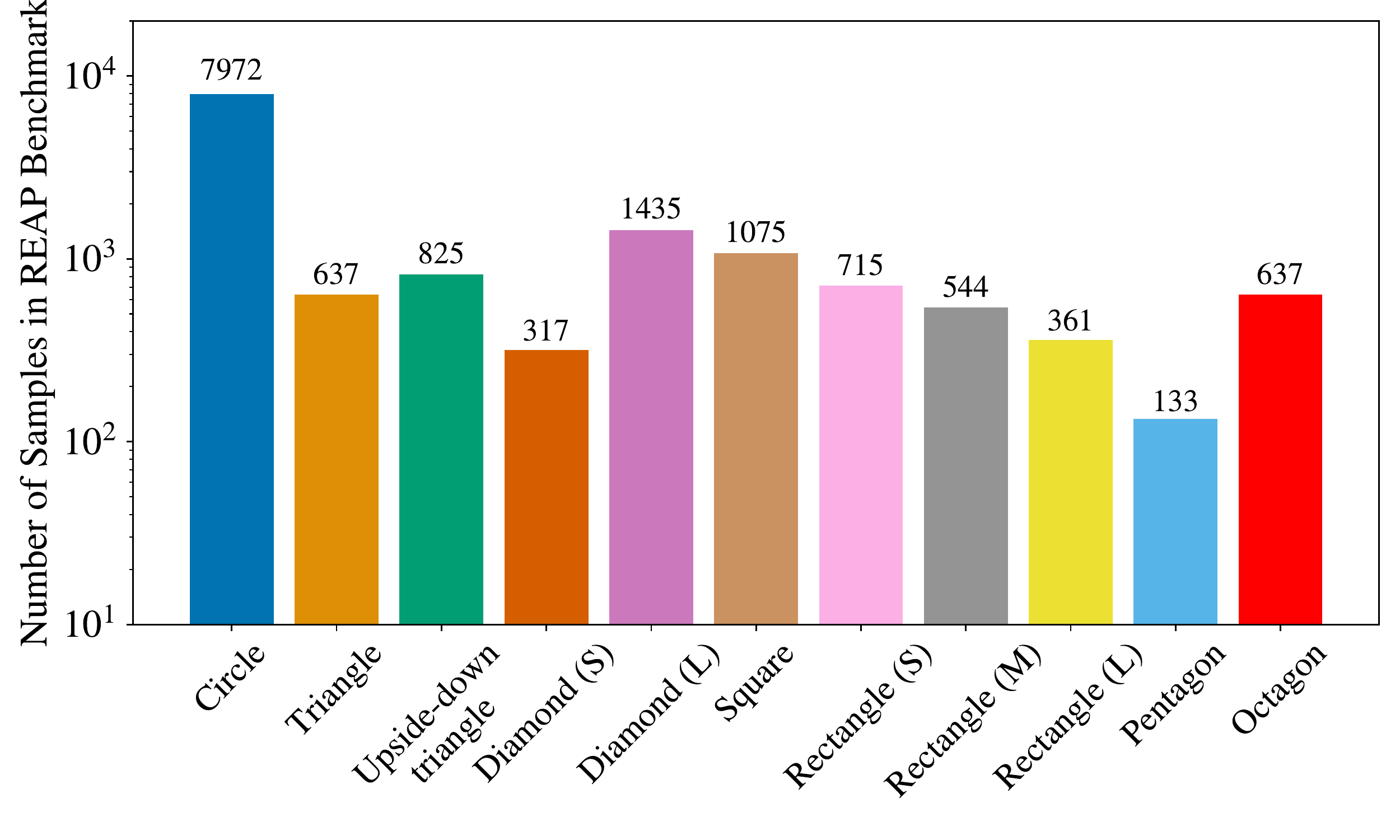}
      \caption{Class distribution of the traffic signs in our \apbs{} benchmark.}\label{fig:class_distribution}
   \end{minipage}
   \hfill
   \begin{minipage}{.38\textwidth}
      \centering
      \includegraphics[width=\textwidth]{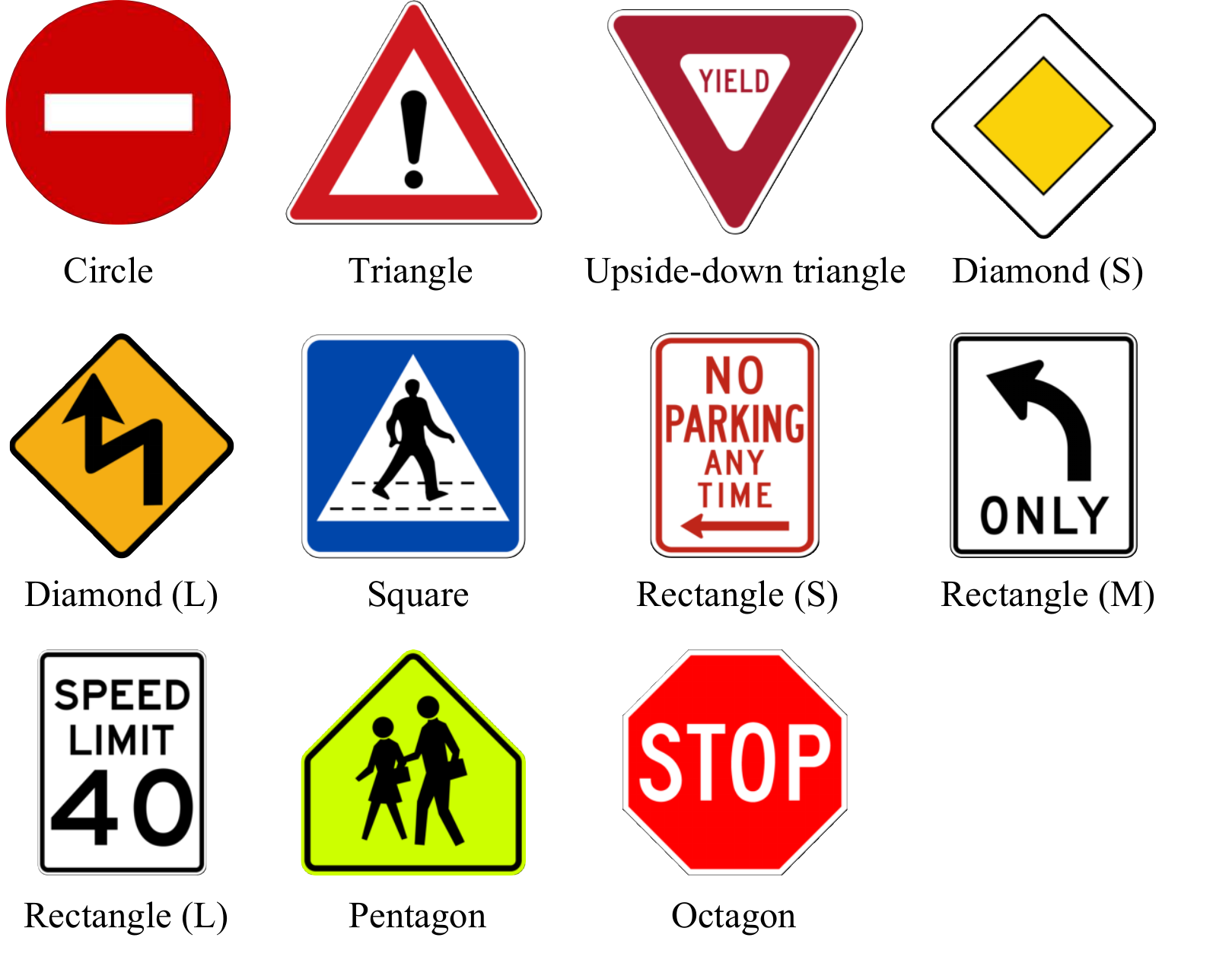}
      \caption{Images of the synthetic signs we use for computing the relighting transform parameters and evaluating adversarial patches under the synthetic benchmark.}\label{fig:synthetic_signs}
   \end{minipage}
\end{figure}

\begin{figure}[t]
   \centering
   \includegraphics[width=0.9\textwidth]{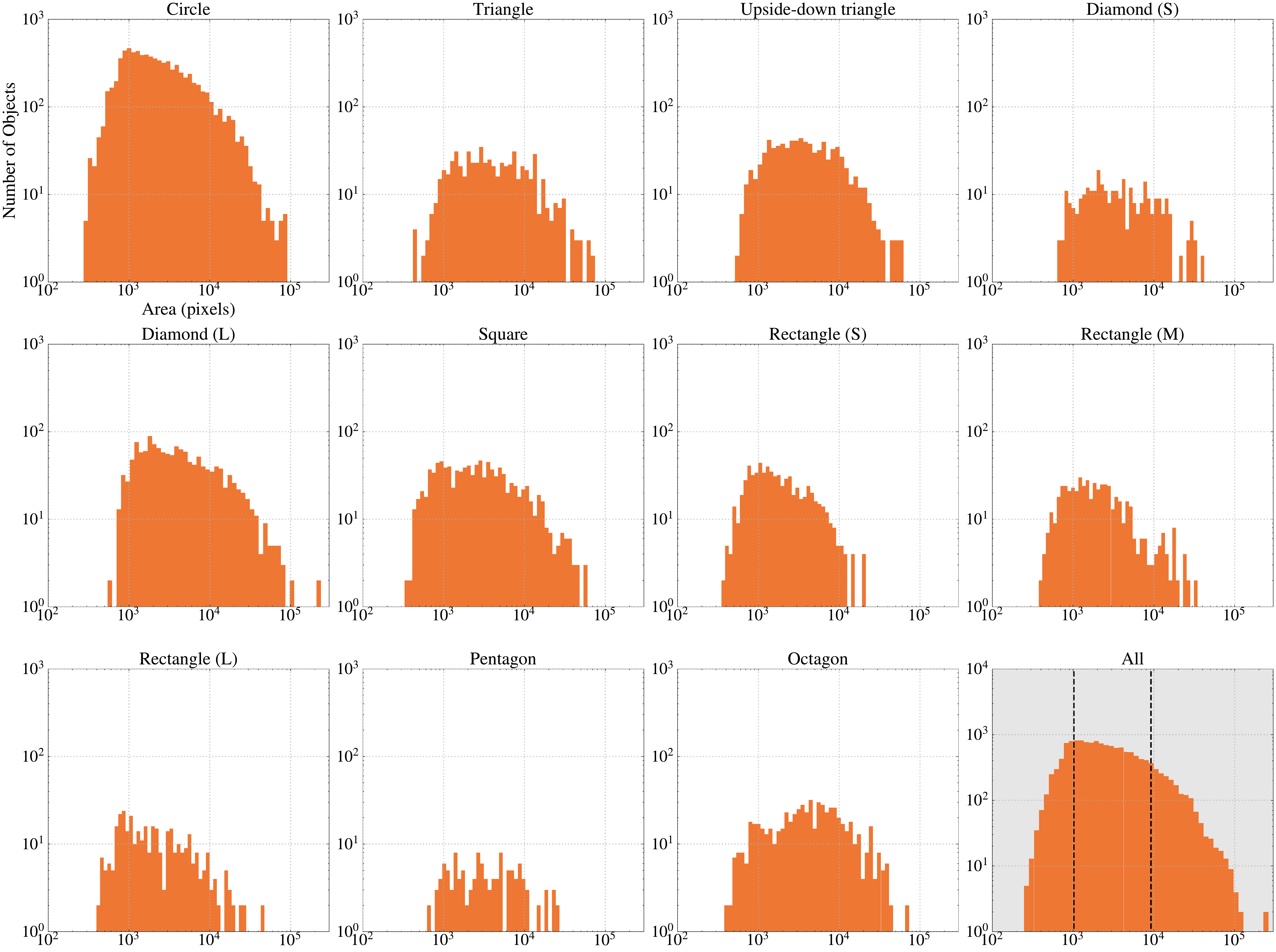}
   \caption{Distribution of the sign sizes in pixels for each class and all combined (the last or bottom right panel). The two black lines in the last panel split the signs into three groups based on the area defined by COCO~\citep{lin_microsoft_2014}: small (--32$\times$32), medium (32$\times$32--96$\times$96), and large (96$\times$96--), from left to right.}\label{fig:stat_obj_sizes}
\end{figure}

\begin{figure}[t]
   \centering
   \begin{subfigure}[b]{0.77\textwidth}
      \centering
      \includegraphics[width=\textwidth]{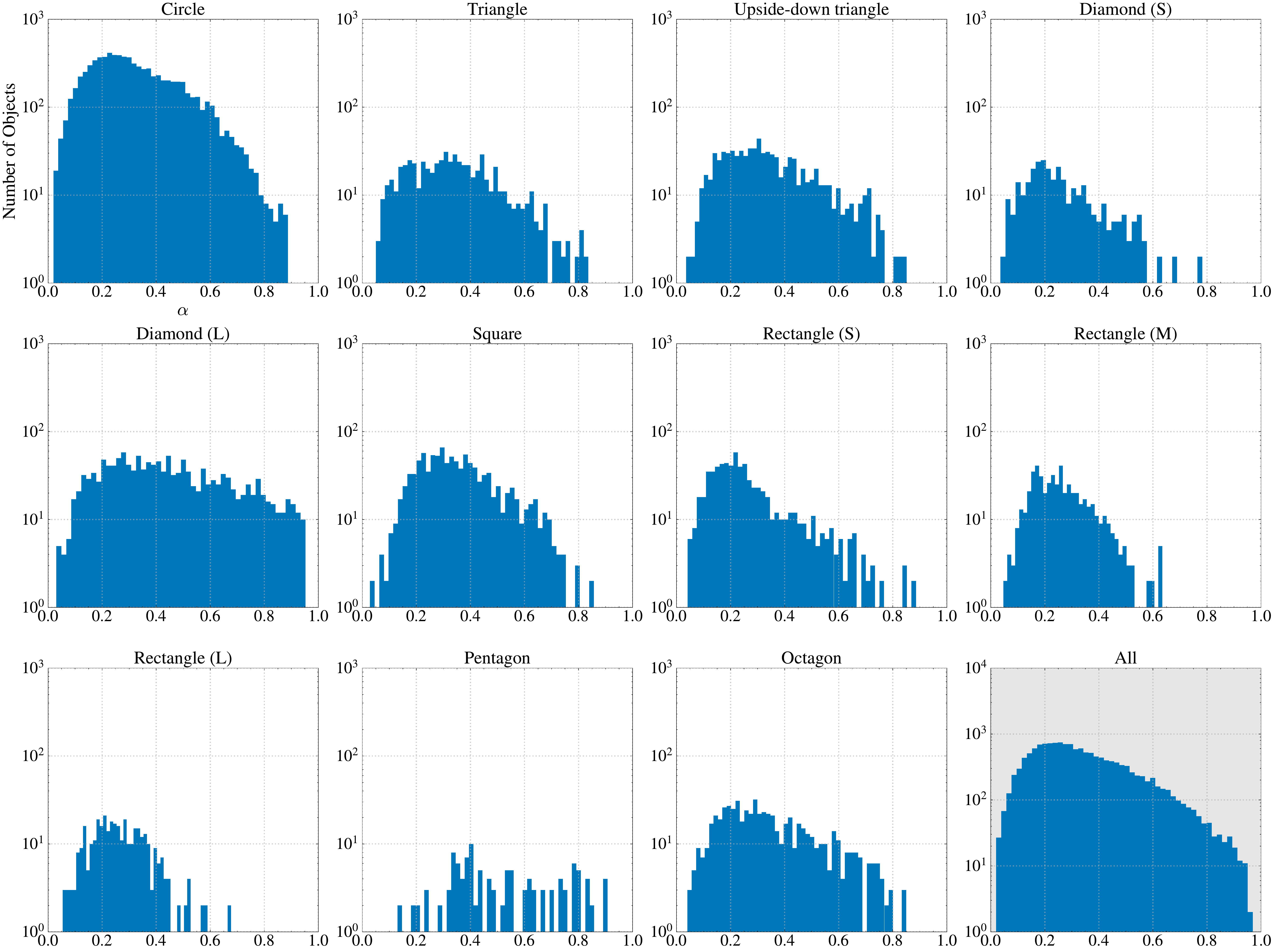}
      \caption{Distribution of the scaling parameter, $\alpha$, from the percentile relighting method.}\label{fig:stat_alphas}
   \end{subfigure}
   \begin{subfigure}[b]{0.77\textwidth}
      \centering
      \includegraphics[width=\textwidth]{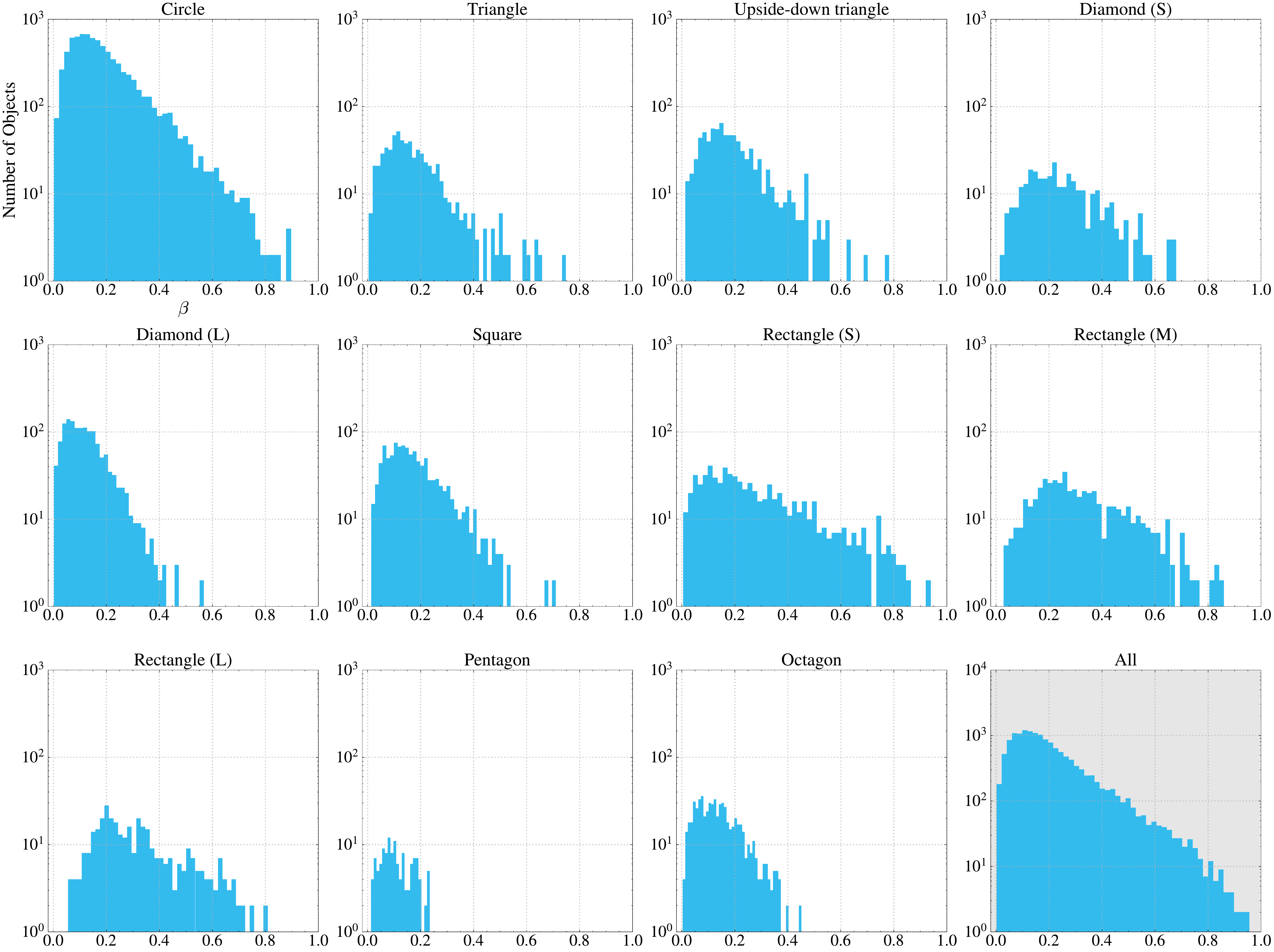}
      \caption{Distribution of the offset parameter, $\beta$, from the percentile relighting method.}\label{fig:stat_betas}
   \end{subfigure}
   \caption{Class-wise distribution of the parameters from the percentile relight method. The last panel with gray background combines all classes.}\label{fig:stat_percentile}
\end{figure}

\paragraph{Dataset Statistics.}
\cref{fig:class_distribution} and \cref{tab:sign_details} summarize the class distribution of the \apbs{}' traffic signs.
\cref{tab:sign_details} also contains the ``standardized'' size in inches for each class of the signs.
\cref{fig:stat_obj_sizes,fig:stat_percentile} show the distribution of the \apb's traffic sign sizes (area in pixels) and the two parameters, $\alpha, \beta$, from the percentile relighting method, respectively.
All distributions differ from class to class.
\cref{fig:synthetic_signs} shows images of the digital signs we use for computing the relighting transform parameters and evaluating adversarial patches under the synthetic benchmark.

\begin{table}[t]
   \caption{Dimension of the sign by classes in the \apbs{} benchmark.}\label{tab:sign_details}
   \vspace{-10pt}
   \small
   \centering
   \begin{tabular}{@{}lrrr@{}}
      \toprule
      Traffic Sign Class   & Width (mm) & Height (mm) & Number of Samples in \apbs{} \\ \midrule
      Circle               & 750        & 750         & 7971                         \\
      Triangle             & 900        & 789         & 636                          \\
      Upside-down triangle & 1220       & 1072        & 824                          \\
      Diamond (S)          & 600        & 600         & 317                          \\
      Diamond (L)          & 915        & 915         & 1435                         \\
      Square               & 600        & 600         & 1075                         \\
      Rectangle (S)        & 458        & 610         & 715                          \\
      Rectangle (M)        & 762        & 915         & 544                          \\
      Rectangle (L)        & 915        & 1220        & 361                          \\
      Pentagon             & 915        & 915         & 133                          \\
      Octagon              & 915        & 915         & 637                          \\ \bottomrule
   \end{tabular}
\end{table}

\clearpage

\subsection{Traffic Sign Classification}\label{ap:ssec:classify}

Our first step is to simplify the attack setting by grouping traffic signs of a similar shape and size together.
The original MTSD dataset has over 300 classes so we hope to take a subset of them with a common and fairly standardized size.
For example, we do not take generic directional signs because there is no standard size for them at all.
In the end, we end up with 11 classes in total plus one background class where all the remaining signs belong.
Then, we assign the dimension to each class according to the official guideline published by the U.S. Department of Transportation as mentioned in \cref{ssec:classify}.
\cref{tab:sign_details} summarizes dimension we assign to all the signs.

This dimension only approximates the true size which we have no way of measuring given that the camera specifications and the distance to the signs are not known.
This leads to two limitations: First, signs within a single class may actually be of different sizes because each sign has more than one standard size which mostly depends on the type of road it is placed on.
The second is that both MTSD and Mapillary Vistas contain signs from all over the world and not only from the US.
Hence, the dimension may not be consistent across countries.
Nevertheless, we argue that this approximation is more realistic than naively specifying the patch size relative to the sign size in pixels (e.g., ``each patch covers 10\% of the sign'') because sign sizes vary significantly between classes.

After mapping the original MTSD classes to our new 11 classes, we train a ConvNeXt-Small~\citep{liu_convnet_2022} on all of the signs from MTSD.
These signs are cropped by leaving 10\% padding between the sign border and the patch border on all four sides.
The cropped patches are then resized to 128$\times$128 pixels.
The model was pre-trained on ImageNet-22k, and we fine-tune it with a batch size of 128, a learning rate of $0.1$, and a weight decay of $5 \times 10^{-4}$.
We use the validation set to early stop the training where the model achieves slightly above 98\% accuracy.
Lastly, we use the trained ConvNeXt to classify traffic signs in the Mapillary dataset.
We also combine the training and the validation sets together.
We ignore signs that are classified as the background class and discard images that do not contain any non-background sign.

After obtaining the pseudo-labels on the Mapillary Vistas dataset, we verify that they are consistent with the shape-based classes of \apbs{} that we have already manually verified.
For ones that are inconsistent, we automatically re-label them using subsequently less predicted classes as long as its confidence score is above a certain threshold.
In the end, we simply drop the samples that we cannot re-label and include them as the ``other'' class instead.
Note that some of the signs in \apbs{} are not included in the 100 most common classes, or incorrectly classified, or unable to be automatically re-labeled.
Consequently, \apb{} ends up with a slightly smaller number of samples than \apbs{} in total.

\section{Realism Test}\label{ap:sec:realism_test}

In this section, we hope to justify our choice of the geometric and the relighting transform methods.
The geometric transform is arguably more straightforward since we should prefer the most flexible (the most degrees of freedom) transform which is the perspective or 3D transform.
Modeling the change in lighting is, on the other hand, more challenging as the true transform function is unknown, let alone its parameters.
Hence, we need an approximation that is sufficiently fast to compute, differentiable, and can be determined easily enough based only on the RGB pixel values.
In this section, we will first introduce all the relight transform methods we consider along with the metric we use to rank them.

\subsection{Relighting Transform Candidates}

The idea is that we want to find some transform $t_\theta(\cdot)$ that maps a digital sign to a real sign.
We will experiment with different classes of the function $t$, and the parameters $\theta$ are determined by the pixel values of the real and the digital signs.
Then, we will apply the same transform to the digital adversarial patch to simulate the lighting condition of that real sign.
One common approach for matching the lighting condition of one image to another is \emph{histogram matching} based on \citet{shapira_multiple_2013}.
However, this method is computationally intensive and is unclear how to  determine the transform on a pair of digital and real signs and apply it to the digital patch.

\paragraph{The polynomial method.}
The next most reasonable approach is to approximate the matching process with some function class that can be applied \emph{pixel-wise}.
We choose a polynomial function of degree $k$ and the parameters $\theta$ are the coefficients of the polynomial, i.e., $\theta = \{\theta_0, \theta_1, \ldots, \theta_k\}$.
The $\theta$'s are then determined by fitting this function on the pixel value of the digital sign to predict the corresponding pixel value on the real sign.
Note that once we have a reasonable geometric transform, we can get a one-to-one mapping between a pixel on the digital sign to a pixel on the real one.
Let all pairs of the digital-real pixel values be $\{(x_{d,i}, x_{r,i})\}_{i=1}^N$ where $N$ is the number of pixels that the real sign occupies in the image (obtained from a transformed mask).
Then, $\{\theta_0, \theta_1, \ldots, \theta_k\}$ are given by
\begin{align}
   \{\theta_0, \theta_1, \ldots, \theta_k\} ~       = ~\argmin_{\{\tilde{\theta}_0, \tilde{\theta}_1, \ldots, \tilde{\theta}_k\} \in \R^{p+1}}~ R\left(\{(x_{d,i}, x_{r,i})\}_{i=1}^N; \theta\right) & \coloneqq \quad \argmin_{\{\tilde{\theta}_0, \tilde{\theta}_1, \ldots, \tilde{\theta}_k\} \in \R^{p+1}}~ \frac{1}{N}\sum_{i=1}^N \left(t_{\tilde{\theta}}(x_{d,i}) - x_{r,i} \right)^2 \\ \label{eq:rmse}
   \text{where} \quad t_{\tilde{\theta}}(x_{d,i})                                                                                                                                                    & = \tilde{\theta}_0 + \tilde{\theta}_1x_{d,i} + \tilde{\theta}_2x_{d,i}^2 + \cdots + \tilde{\theta}_k x_{d,i}^k
\end{align}
We call this method ``polynomial'' and experiment with $k \in \{0,1,2,3\}$.
Additionally, instead of just computing the error $R\left(\{(x_{d,i}, x_{r,i})\}_{i=1}^N; \theta\right)$ as MSE, we also experiment with trimmed mean by cutting off both tails at $p$ and $1-p$ percentiles.
This is intended to reduce the effect of outliers or noisy pixels.

\paragraph{Percentile method.}
The percentile method is a slightly simplified version of the polynomial method.
In particular, we restrict the transform $t_\theta(\cdot)$ to be a linear function of the pixel value, i.e., $t_\theta(x) = \theta_0 + \theta_1 x$.
However, instead of simply matching the pairs of the pixels, we view this problem as an affine scaling that simply matches the minimum and the maximum of the pixel values.
For digital signs, the minimum and the maximum are always 0 and 255 (or the scaled 0 and 1), respectively, but on the real signs, they almost always lie in a smaller range.
The minimums and the maximums can then be matched and fully determined by an affine transform (2 parameters and 2 unknowns).
We call $\theta_1$ and $\theta_0$ as $\alpha$ and $\beta$ instead to differentiate it from the polynomial method, and similarly, we use the $p$-th and $(1-p)$-th percentiles, in place of max and min, to mitigate the effect of outliers.

\future{give formula}

\paragraph{Color transfer method.}
The color transfer method is proposed \citet{reinhard_color_2001} to process the color of one image to match that of a reference one.
This is achieved by matching the mean and the standard deviation of the pixel values in the LAB color space.
To match the lighting of the digital patch to the real one, we adapt this method to only consider the channel that deals with lighting, i.e., the L channel, and ignore the other two channels (A and B channels).
Keeping the other two channels ends up changing the color of the patch drastically and yields a worse result visually.

\paragraph{Choices of Color Spaces.}
One important consideration is that we cannot apply this transform independently on the three RGB channels.
Otherwise, it will also change the color or the hue of the patch completely.
To mitigate this problem, we choose to apply the polynomial method and determine its parameters in three different ways: (1) on the \emph{maximum} value among the RGB channels, (2) on the S and V channels of HSV color space, and (3) on the L channel of LAB color space.

\subsection{Metric and Results}

\begin{figure}[t]
   \centering
   ~\hfill
   \begin{subfigure}[b]{0.33\textwidth}
      \includegraphics[width=\linewidth]{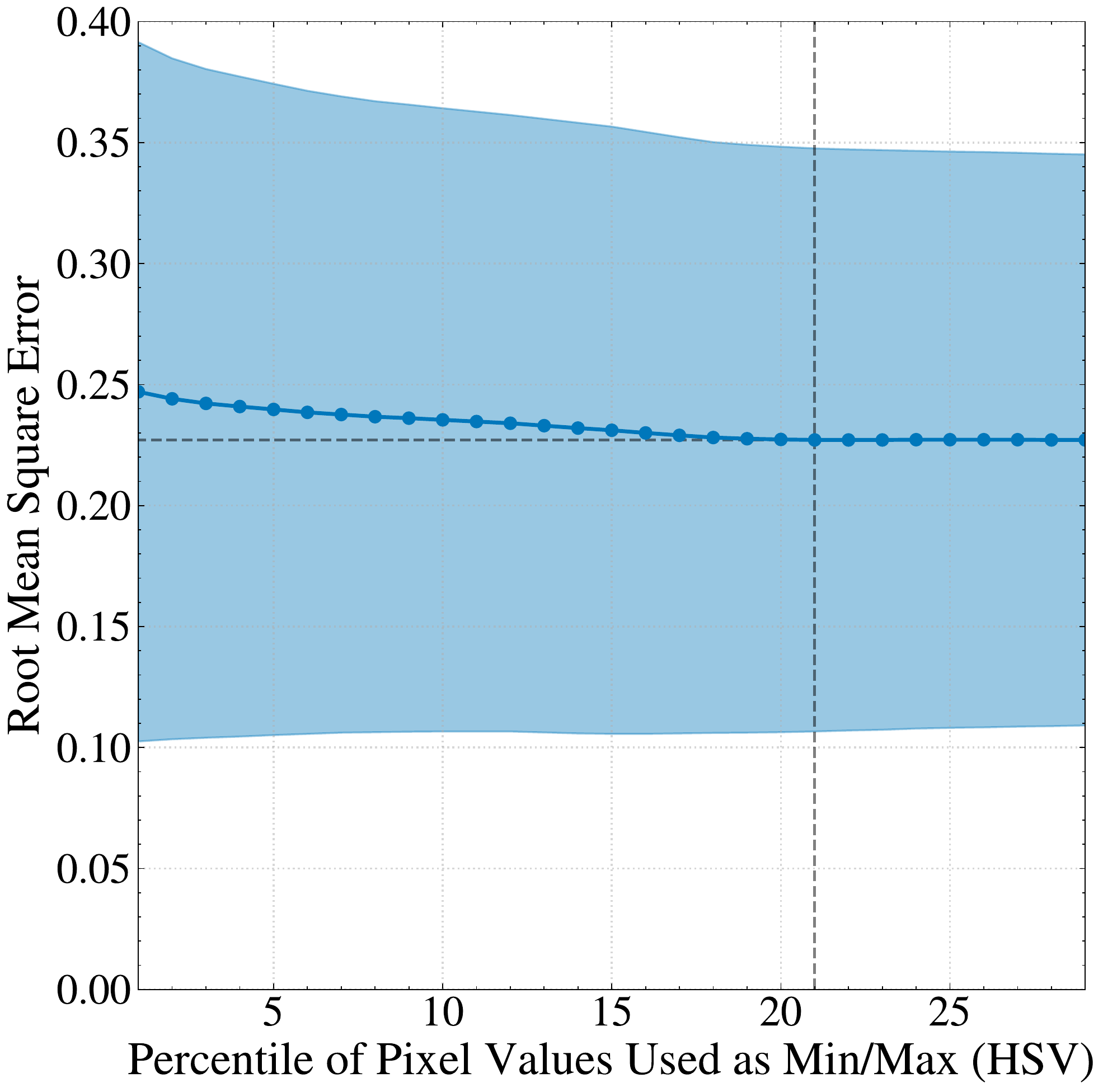}
      \caption{HSV}
   \end{subfigure}
   \hfill
   \begin{subfigure}[b]{0.33\textwidth}
      \includegraphics[width=\linewidth]{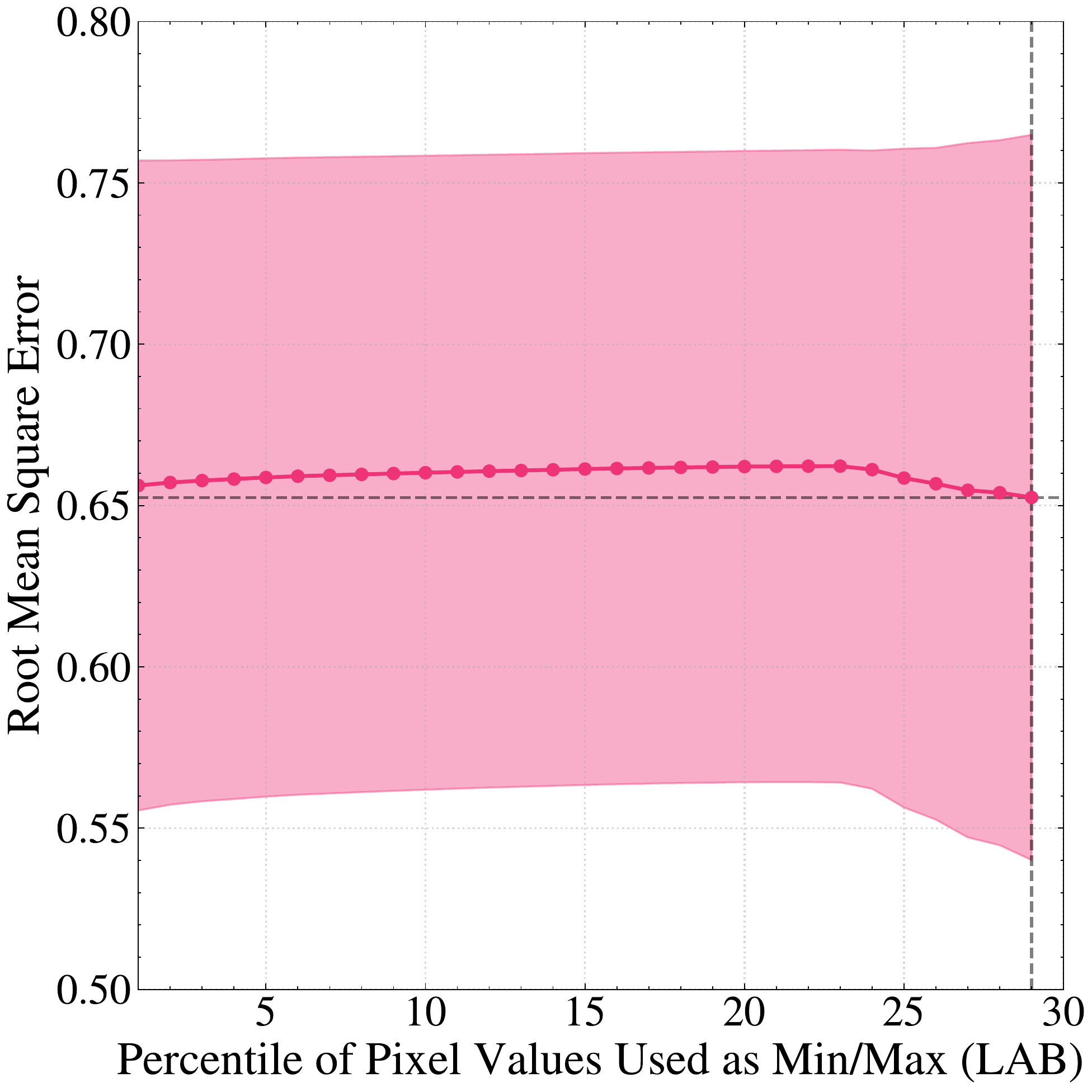}
      \caption{LAB}
   \end{subfigure}
   \hfill~
   \caption{RMSE between the real and the rendered patches averaged across all samples from our realism test dataset. The rendering is done by the \emph{percentile} method under (a) HSV and (b) LAB color spaces. The RGB color space is shown in \cref{fig:realism_test_percentile_rgb}.}\label{fig:realism_test_percentile}
\end{figure}

\begin{figure}[t]
   \centering
   \begin{subfigure}[b]{0.32\textwidth}
      \includegraphics[width=\linewidth]{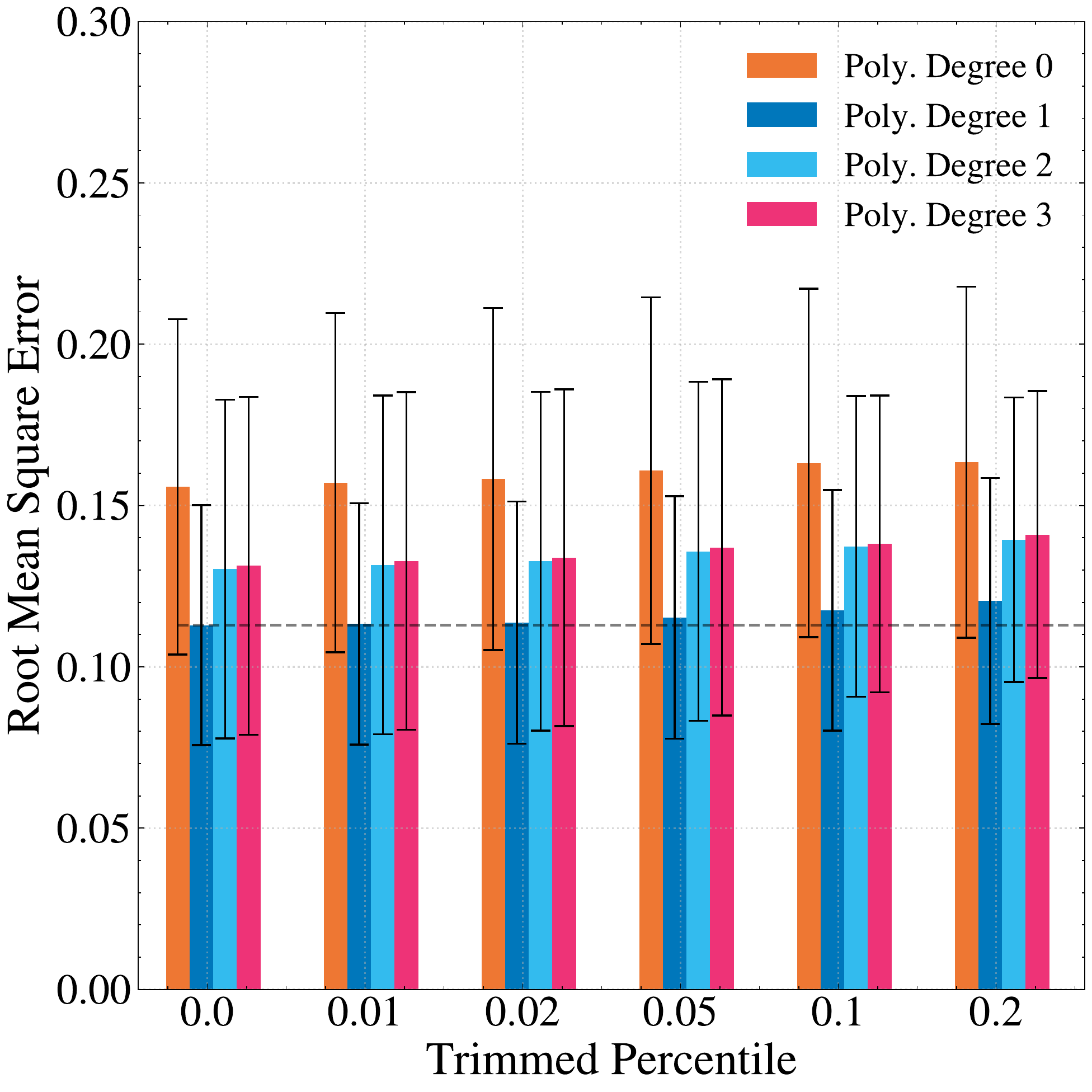}
      \caption{RGB}
   \end{subfigure}
   \hfill
   \begin{subfigure}[b]{0.32\textwidth}
      \includegraphics[width=\linewidth]{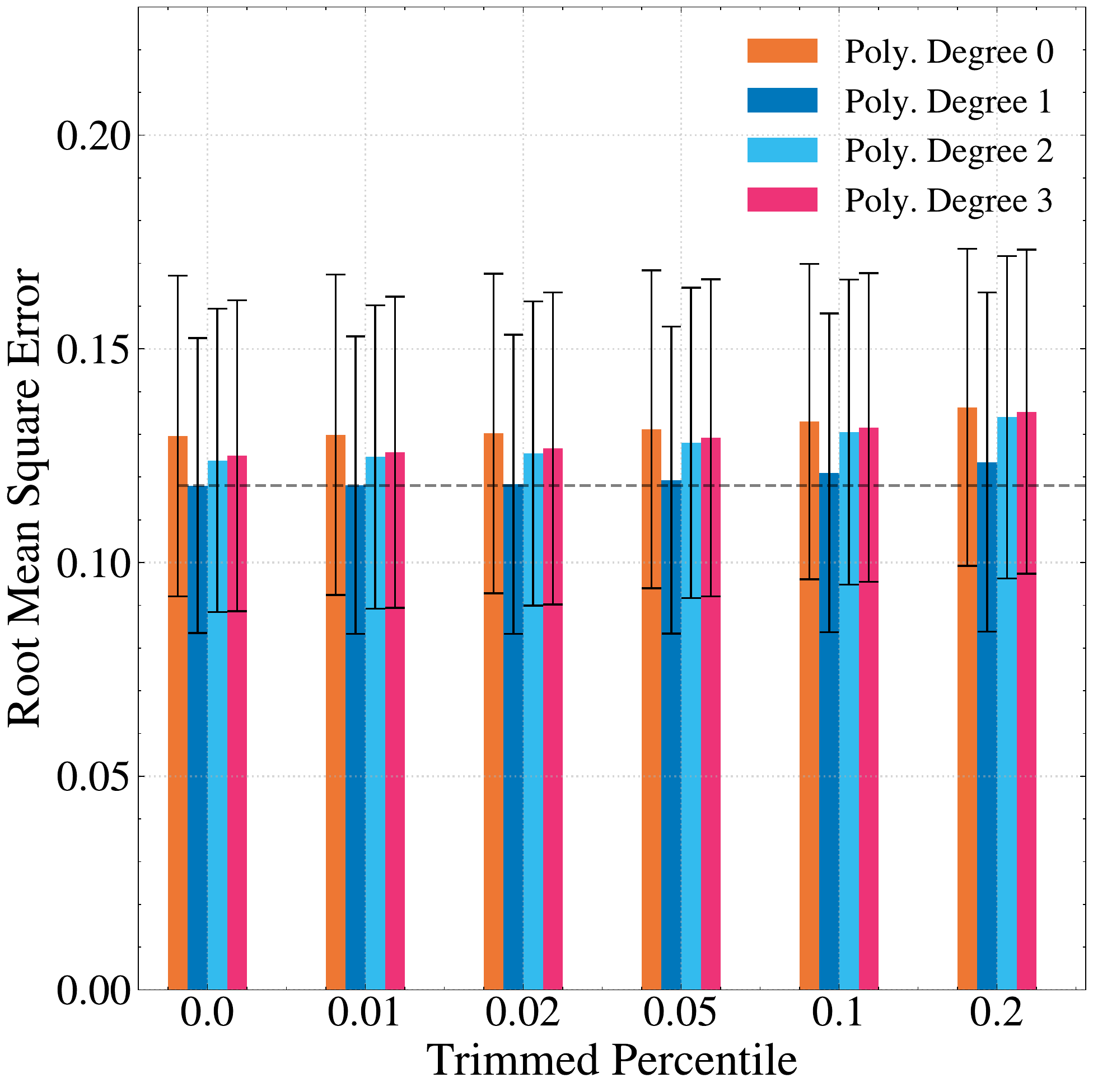}
      \caption{HSV}
   \end{subfigure}
   \hfill
   \begin{subfigure}[b]{0.32\textwidth}
      \includegraphics[width=\linewidth]{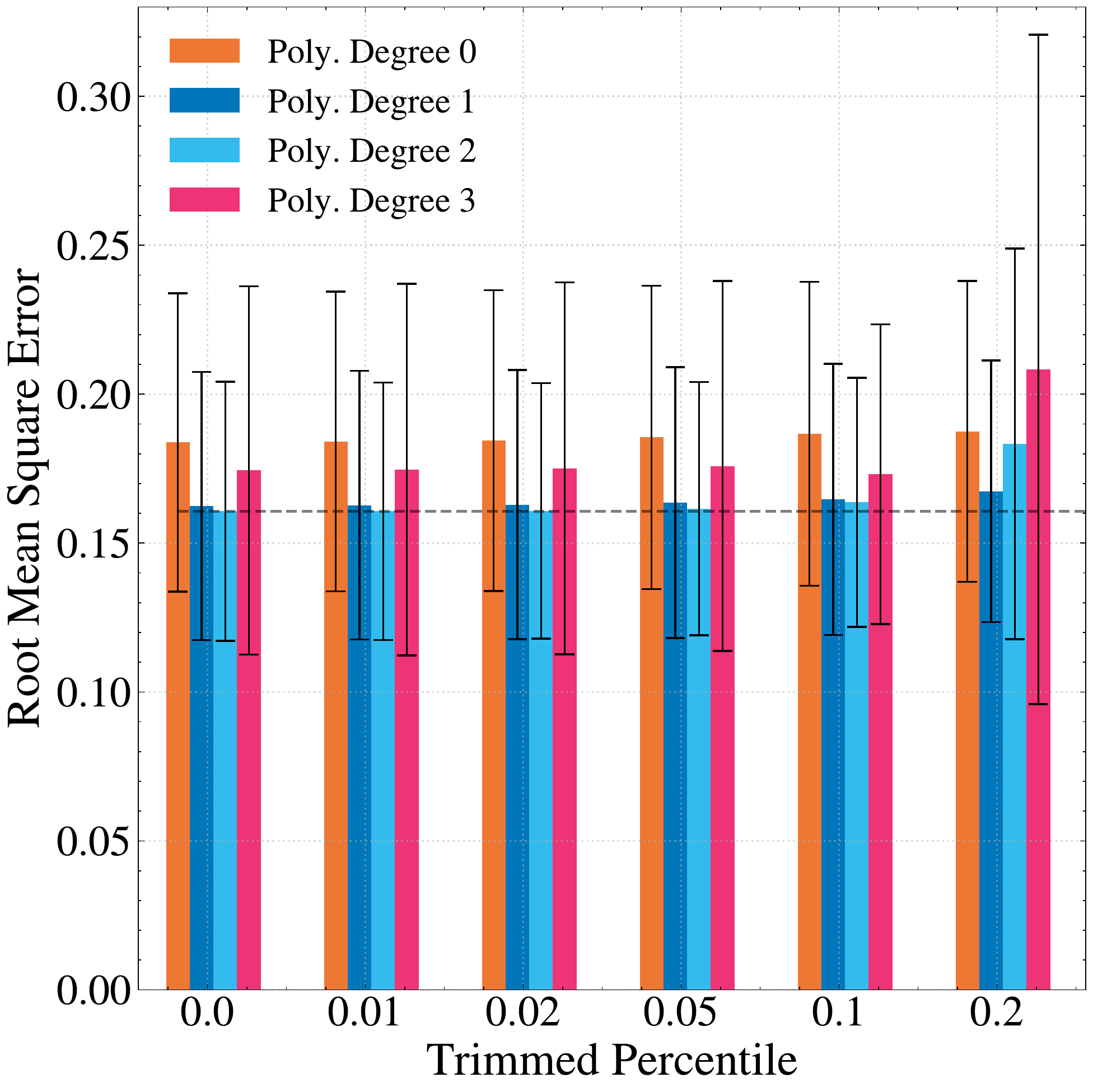}
      \caption{LAB}
   \end{subfigure}
   \caption{RMSE between the real and the rendered patches averaged across all samples from our realism test dataset. The error bar denotes one standard deviation. The rendering is done by the \emph{polynomial} method under (a) RGB, (b) HSV, and (c) LAB color spaces. We sweep over the choices of polynomial degree $k \in \{0,1,2,3\}$ and the trimmed percentile $p \in \{0, 0.01, 0.02, 0.05, 0.1, 0.2\}$. The horizontal dashed lines denote the lowest RMSE across all the parameter choices for each color space respectively}\label{fig:realism_test_polynomial}
\end{figure}

\begin{figure}[t]
   \centering
   \includegraphics[width=0.85\linewidth]{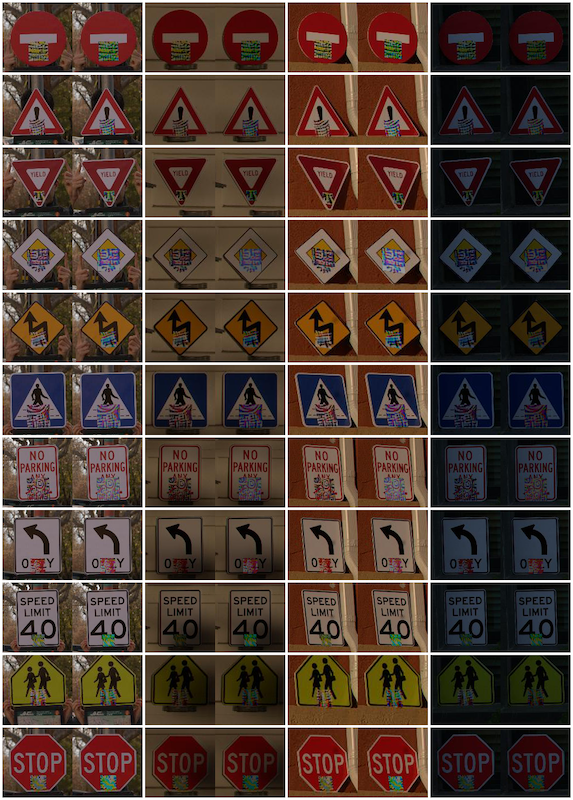}
   \caption{All samples from the realism test (left: real, right: rendered). Each column of images is taken in the same scene and lighting condition. Each row is the same sign and the same patch. The patches (right) are rendered with the ``percentile'' relighting with $p=0.2$ and the 3D perspective transform.}\label{fig:realism_test_all}
\end{figure}

To compare the geometric transform method, we also compute the root mean squared error (RMSE) between the corners of the real and the rendered patches, instead of the pixel values.
Since it does not make sense to compare the transformed coordinates through pixel values, we choose to compare the (square) Euclidean distance between the groundtruth and the corresponding rendered points in the 2D coordinate space.

To compare the accuracy of the three relighting methods with various parameters, we use the RMSE between the real and the rendered patches.
This metric is computed in the exact same way as the objective function in \cref{eq:rmse}.
We emphasize that the patches are only compared after the rendered one is geometrically transformed to be in the exact same orientation as the real counterpart.
This transform is not inferred from the keypoints of the traffic signs as done in the generation of the \apb{} benchmark, but it is computed from four corners of the patches each of which is hand-labeled.
This essentially separates out the choice (and potentially the error) of the geometric transform from the relighting method.

\cref{tab:realism_main} summarizes the results for all the geometric and the relighting transforms according to the metric described above.
Due to the space limit, we only report the \emph{best} results for the percentile and the polynomial methods across the parameter sweep.
The full results of these two methods are included in \cref{fig:realism_test_percentile_rgb,fig:realism_test_percentile,fig:realism_test_polynomial}.

\clearpage

\section{Detailed Experiment Setup}\label{ap:sec:exp_setup}

\subsection{Object Detection Models}\label{ap:ssec:models}

All the object detection models, Faster R-CNN~\citep{ren_faster_2015}, YOLOF~\citep{chen_you_2021}, and DINO~\citep{zhang_dino_2022}, are trained on the MTSD dataset with an input size of $1536 \times 2048$.
Both the Faster R-CNN and the YOLOF models use ResNet-50 backbone (pre-trained on ImageNet-1k), and the DINO model uses the Swin-Tiny backbone (pre-trained on ImageNet-22k)~\citep{liu_swin_2021}.
We do not use any data augmentation as the models trained on the clean data already perform well without.
It is likely that adding data augmentation will improve their performance as well as robustness further.
We use the default hyperparameters provided by the authors for all the models.

To compute the FNR and the ASR metric, we pick a specific confidence score threshold that maximizes the F1 score on the clean \apb{} samples for each model and each class.
For the synthetic data, we have to choose a new score threshold since the data distribution is different.
Here, we choose the threshold that results in the same FNR as that of the \apb{} dataset which allows us to fairly compare the FNR as well as the ASR metrics across these two datasets.
We use the same threshold when evaluating against the patch attacks.

\subsection{Patch Attack Algorithms}\label{ap:ssec:attacks}

We re-implement both the RP2~\citep{eykholt_physical_2018} and the DPatch~\citep{liu_dpatch_2019,lee_physical_2019} attacks based on the description provided in the paper.
The main difference between these two attacks is that RP2 focuses on making an untargeted misprediction on the objects while DPatch optimizes over the same loss as the training loss so it additionally affects the objectness score as well as the predicted bounding boxes.

For both attacks, our default hyperparameters include 64$\times$64 patch and object dimension, EoT rotation with a maximum 15 degrees, no color jitter for EoT.
We set the step size to 0.1 and 0.01 when using Adam and PGD optimizers, respectively.
In most scenarios, the attack is run for 1,000 iterations which are more than enough to converge.
For per-instance attacks, we use a PGD step size of 0.02 and run for 100 iterations instead, due to the much higher computation requirement; this still takes about 1 GPU week.

The $\lambda$ parameter used to encourage low-frequency patterns is set to $10^{-5}$.
We find that the choices of the patch dimension and $\lambda$ do not affect the ASR when varying in a reasonable range.
We use the same set of hyperparameters when generating adversarial patches from both our \apb{} and the synthetic benchmark.

\section{Additional Experiments}\label{ap:sec:more_exp}

\subsection{Comparison of Attack Algorithms and Optimizers}\label{ap:ssec:rp2_vs_dpatch}

\begin{table}[t]
    \centering
    \small
    \caption{Attack success rates by sign classes of four combinations of attack algorithms (RP2 and DPatch) and optimizers (Adam, PGD) on \apbs{} benchmark. The patch size is $\inch{10} \times \inch{10}$.}\label{tab:rp2_vs_dpatch}
    {
        \setlength\tabcolsep{5.0pt}
        \begin{tabular}{ll|rrrrrrrrrrr|r}
            \toprule
            Models & Attacks       & Circ & Tri  & UTri & Dia(S) & Dia(L) & Squ  & Rec(S) & Rec(M) & Rec(L) & Pen  & Oct  & \textit{Avg.} \\ \midrule
            \multirowcell{4}[0pt][l]{Faster R-CNN}
                   & RP2 (Adam)    & 28.9 & 62.3 & 3.4  & 93.3   & 6.3    & 73.6 & 62.9   & 33.7   & 13.7   & 17.8 & 33.5 & 39.0          \\
                   & RP2 (PGD)     & 30.9 & 61.9 & 3.1  & 91.0   & 6.7    & 78.5 & 63.2   & 31.0   & 11.2   & 15.1 & 24.4 & 37.9          \\
                   & DPatch (Adam) & 30.3 & 64.6 & 3.4  & 87.6   & 6.3    & 76.7 & 64.3   & 35.3   & 13.7   & 42.5 & 28.3 & 41.2          \\
                   & DPatch (PGD)  & 26.2 & 65.7 & 3.8  & 89.1   & 6.1    & 80.1 & 62.0   & 34.0   & 10.2   & 26.0 & 27.7 & 39.2          \\
            \cmidrule{2-14}
            \multirow{4}{*}{YOLOF}
                   & RP2 (Adam)    & 29.5 & 68.3 & 3.4  & 91.8   & 6.0    & 80.9 & 81.1   & 41.6   & 16.7   & 81.4 & 27.5 & 48.0          \\
                   & RP2 (PGD)     & 26.9 & 63.9 & 3.2  & 90.4   & 5.8    & 75.6 & 79.5   & 29.7   & 13.1   & 84.3 & 27.1 & 45.4          \\
                   & DPatch (Adam) & 27.5 & 62.9 & 3.1  & 88.6   & 4.5    & 65.9 & 77.2   & 30.2   & 15.8   & 77.1 & 23.2 & 43.3          \\
                   & DPatch (PGD)  & 25.4 & 58.7 & 3.5  & 89.0   & 4.0    & 65.6 & 80.1   & 29.5   & 11.7   & 41.4 & 18.6 & 38.9          \\
            \cmidrule{2-14}
            \multirow{4}{*}{DINO}
                   & RP2 (Adam)    & 22.4 & 13.8 & 1.9  & 78.7   & 5.8    & 92.3 & 65.5   & 36.6   & 3.3    & 20.8 & 2.3  & 31.2          \\
                   & RP2 (PGD)     & 21.7 & 14.2 & 2.6  & 79.5   & 5.6    & 91.9 & 67.5   & 37.9   & 2.5    & 22.2 & 2.3  & 31.6          \\
                   & DPatch (Adam) & 21.8 & 10.9 & 2.3  & 60.6   & 4.9    & 93.0 & 65.9   & 41.4   & 1.7    & 18.1 & 1.8  & 29.3          \\
                   & DPatch (PGD)  & 23.8 & 11.1 & 2.1  & 61.0   & 4.4    & 92.3 & 49.5   & 41.4   & 2.9    & 18.1 & 2.0  & 28.0          \\
            \bottomrule
        \end{tabular}
    }
\end{table}
\begin{table}[t]
    \centering
    \small
    \caption{Attack success rates by sign classes of four combinations of attack algorithms (RP2 and DPatch) and optimizers (Adam, PGD) on \apbs{} benchmark. The models include two adversarially trained Faster R-CNNs, one with RP2 adversarial patch and the other with DPatch, and one adversarially trained YOLOF with DPatch. The patch size is $\inch{10} \times \inch{10}$.}\label{tab:rp2_vs_dpatch_adv}
    {
        \setlength\tabcolsep{5.0pt}
        \begin{tabular}{ll|rrrrrrrrrrr|r}
            \toprule
            Models & Attacks       & Circ & Tri  & UTri & Dia(S) & Dia(L) & Squ  & Rec(S) & Rec(M) & Rec(L) & Pen  & Oct & \textit{Avg.} \\ \midrule
            \multirowcell{4}[0pt][l]{Adv. FRCNN                                                                                          \\(DPatch)}
                   & RP2 (Adam)    & 0.6  & 0.8  & 0.6  & 1.4    & 0.5    & 1.2  & 2.6    & 7.8    & 7.4    & 1.4  & 1.3 & 2.3           \\
                   & RP2 (PGD)     & 0.6  & 0.4  & 0.9  & 0.9    & 0.3    & 0.6  & 1.5    & 7.0    & 5.4    & 1.4  & 1.1 & 1.8           \\
                   & DPatch (Adam) & 0.6  & 0.8  & 1.1  & 2.7    & 0.8    & 7.1  & 14.0   & 11.1   & 7.4    & 0.0  & 0.9 & 4.2           \\
                   & DPatch (PGD)  & 1.2  & 1.2  & 1.4  & 3.6    & 0.9    & 15.9 & 14.3   & 9.6    & 4.9    & 1.4  & 2.0 & 5.1           \\
            \cmidrule{2-14}
            \multirowcell{4}[0pt][l]{Adv. FRCNN                                                                                          \\(RP2)}
                   & RP2 (Adam)    & 0.3  & 1.2  & 1.5  & 1.0    & 0.2    & 0.1  & 19.5   & 6.4    & 5.4    & 0.0  & 1.3 & 3.4           \\
                   & RP2 (PGD)     & 0.6  & 1.2  & 1.9  & 1.0    & 0.6    & 6.5  & 31.2   & 6.9    & 5.9    & 0.0  & 1.7 & 5.2           \\
                   & DPatch (Adam) & 1.4  & 2.3  & 2.0  & 9.1    & 1.1    & 22.3 & 37.3   & 8.3    & 5.4    & 0.0  & 3.3 & 8.4           \\
                   & DPatch (PGD)  & 1.3  & 2.9  & 1.9  & 7.2    & 1.2    & 10.7 & 37.0   & 8.8    & 6.9    & 1.4  & 2.8 & 7.5           \\
            \cmidrule{2-14}
            \multirowcell{3}[0pt][l]{Adv. YOLOF                                                                                          \\(DPatch)}
                   & RP2 (Adam)    & 3.6  & 18.9 & 3.0  & 34.1   & 4.4    & 33.8 & 26.3   & 20.4   & 7.5    & 40.6 & 2.4 & 17.7          \\
                   & DPatch (Adam) & 2.4  & 5.9  & 3.1  & 32.3   & 2.4    & 2.8  & 17.1   & 11.6   & 3.3    & 10.1 & 1.7 & 8.4           \\
                   & DPatch (PGD)  & 4.6  & 11.4 & 3.1  & 31.8   & 3.5    & 20.1 & 21.8   & 22.6   & 5.6    & 13.0 & 1.7 & 12.7          \\
            \bottomrule
        \end{tabular}
    }
\end{table}

\cref{tab:rp2_vs_dpatch,tab:rp2_vs_dpatch_adv} includes the attack success rate of both the RP2 and the DPatch attacks on the \apbs{} benchmark.
Each attack is also used with two different optimizers, Adam~\citep{kingma_adam_2015} and projected gradient descent (PGD).
DPatch with PGD performs consistently well across the adversarially trained models including ones that are trained against DPatch with PGD itself.
So we choose this attack setup for most of our experiments.

\begin{table}[t]
    \centering
    \small
    \caption{Attack success rates by sign classes under synthetic vs the \apbs{} benchmarks. The patch size is $\inch{10} \times \inch{10}$.}\label{tab:10x10_syn_vs_real_all_classes}
    {
        \setlength\tabcolsep{5.0pt}
        \begin{tabular}{ll|rrrrrrrrrrr|r}
            \toprule
            Models & Benchmarks & Circ  & Tri   & UTri & Dia(S) & Dia(L) & Squ   & Rec(S) & Rec(M) & Rec(L) & Pen   & Oct  & \textit{Avg.} \\ \midrule
            \multirowcell{2}[0pt][l]{Faster R-CNN}
                   & Synthetic  & 98.8  & 99.5  & 16.4 & 99.9   & 17.0   & 100.0 & 100.0  & 50.8   & 84.8   & 39.1  & 97.5 & 73.1          \\
                   & \apbs{}    & 26.2  & 65.7  & 3.8  & 89.1   & 6.1    & 80.1  & 62.0   & 34.0   & 10.2   & 26.0  & 27.7 & 39.2          \\ \cmidrule{2-14}
            \multirow{2}{*}{YOLOF}
                   & Synthetic  & 100.0 & 100.0 & 17.1 & 100.0  & 70.8   & 100.0 & 100.0  & 100.0  & 99.9   & 100.0 & 86.5 & 88.6          \\
                   & \apbs{}    & 25.4  & 58.7  & 3.5  & 89.0   & 4.0    & 65.6  & 80.1   & 29.5   & 11.7   & 41.4  & 18.6 & 38.9          \\
            \bottomrule
        \end{tabular}
    }
\end{table}

\begin{figure}[t]
   \centering
   \begin{subfigure}[b]{0.56\textwidth}
      \centering
      \includegraphics[width=\textwidth]{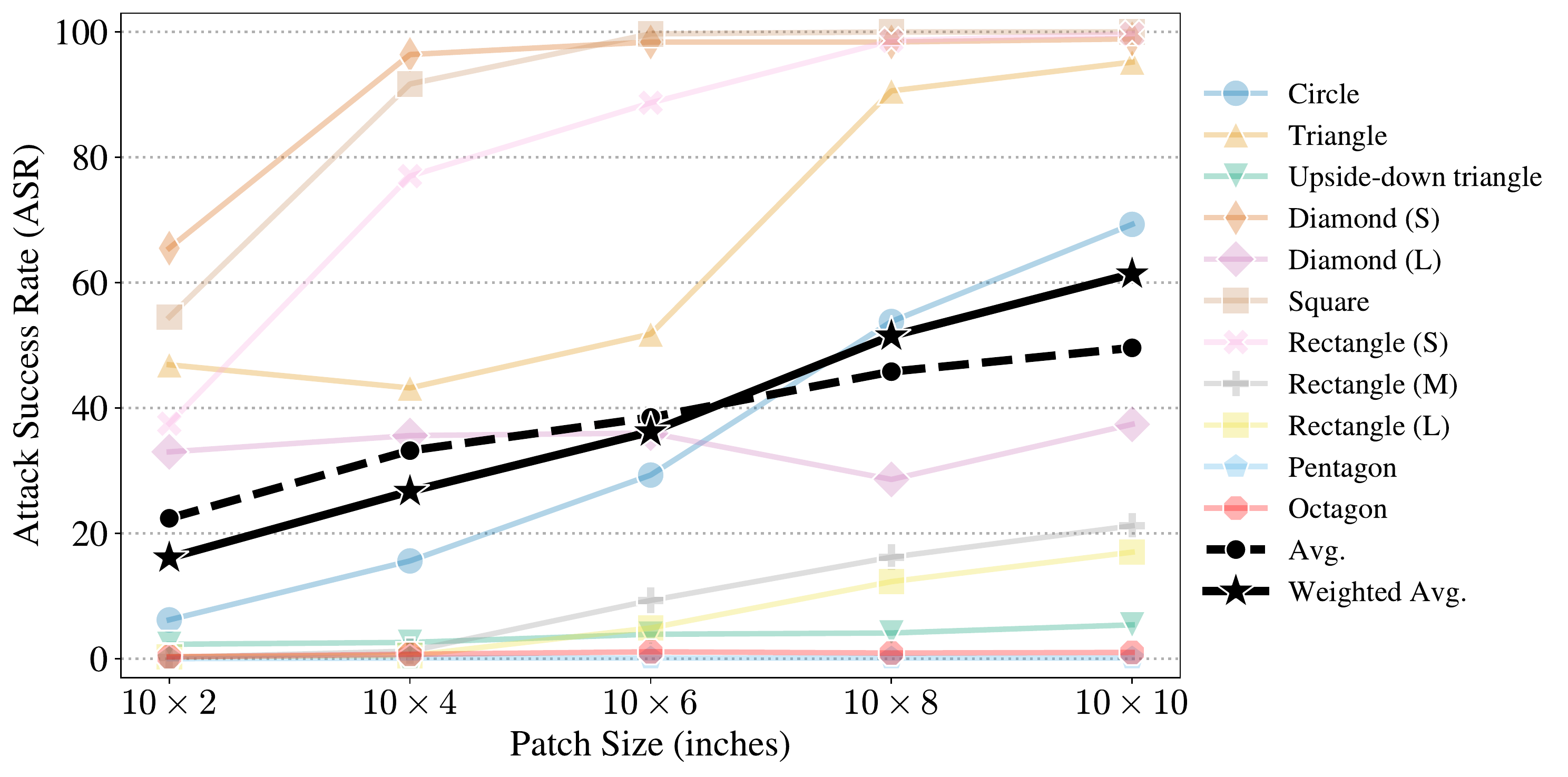}
      \caption{Synthetic benchmark}\label{fig:syn_patch_width}
   \end{subfigure}
   \hfill
   \begin{subfigure}[b]{0.42\linewidth}
      \centering
      \includegraphics[width=\textwidth]{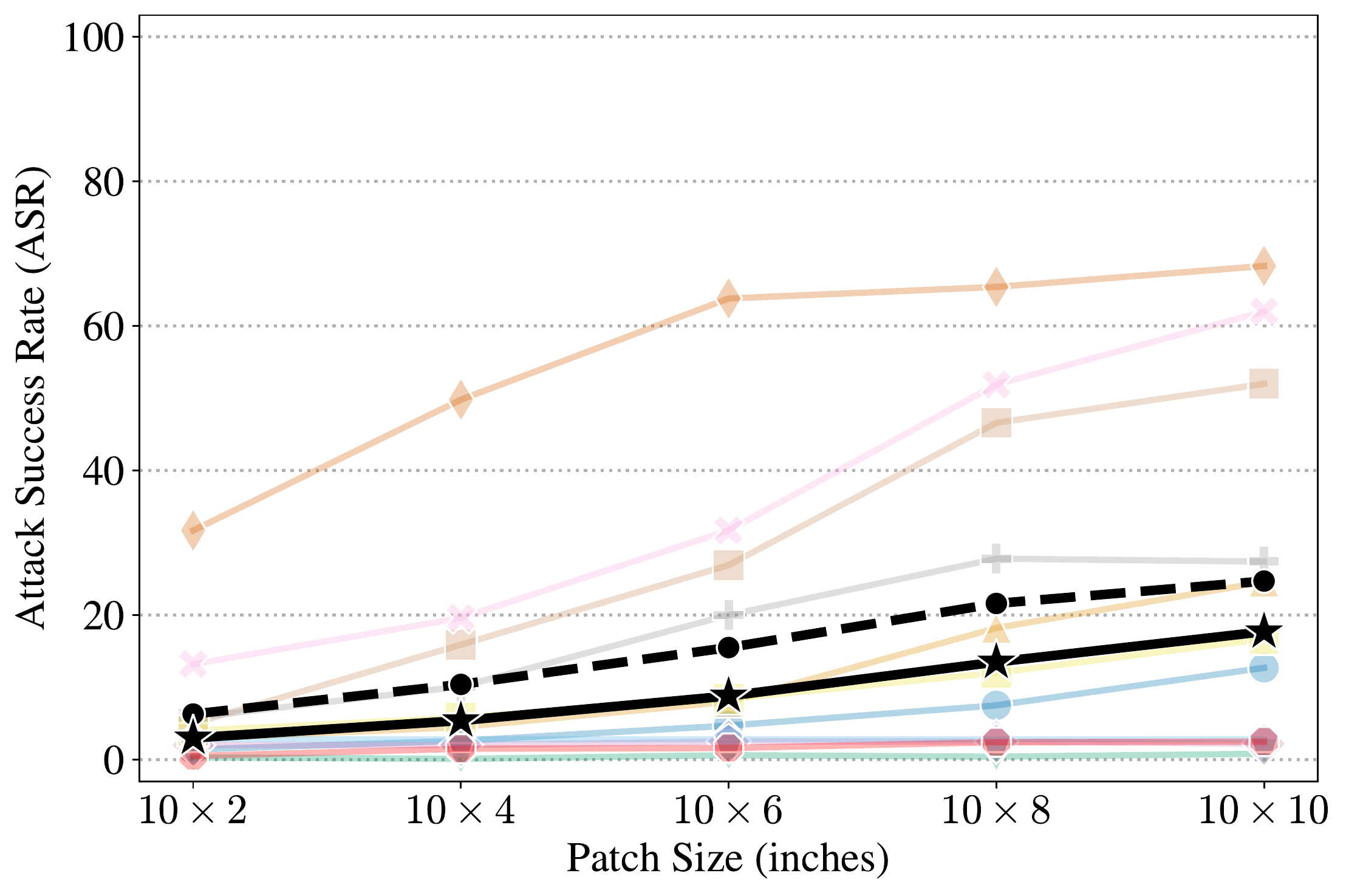}
      \caption{\apbs{} benchmark}\label{fig:real_patch_width}
   \end{subfigure}
   \caption{ASR of all sign classes at different patch sizes on Faster R-CNN. ASR is higher on the synthetic benchmark than our benchmark, for all patch sizes and almost every traffic sign class.}\label{fig:patch_width}
\end{figure}

\subsection{ASR by Classes for Smaller Patch Sizes}

\cref{fig:patch_width} contain a breakdown of ASR by sign class for the patch sizes smaller than \psize{10}{10}.
It is evident that the synthetic data not only overestimate the ASR on average but consistently across almost every traffic sign class.
The trend is also consistent for all patch sizes.

\begin{figure}[t]
   \centering
   ~\hfill
   \begin{subfigure}[b]{0.32\textwidth}
      \centering
      \includegraphics[width=\textwidth]{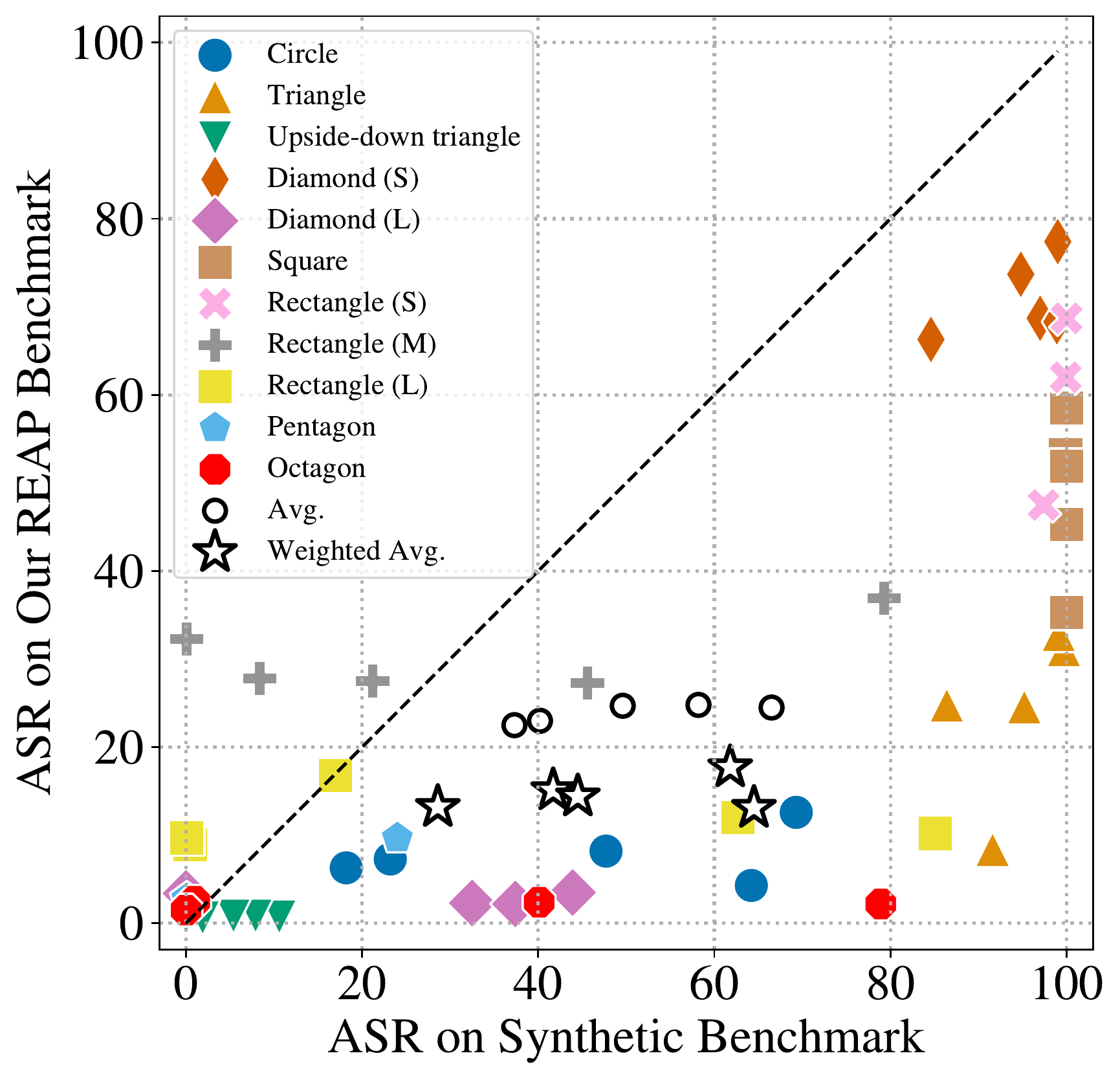}
      \caption{Synthetic object size}
      \label{fig:scatter_10x10_frcnn_obj_size}
   \end{subfigure}
   \hfill
   \begin{subfigure}[b]{0.32\textwidth}
      \centering
      \includegraphics[width=\textwidth]{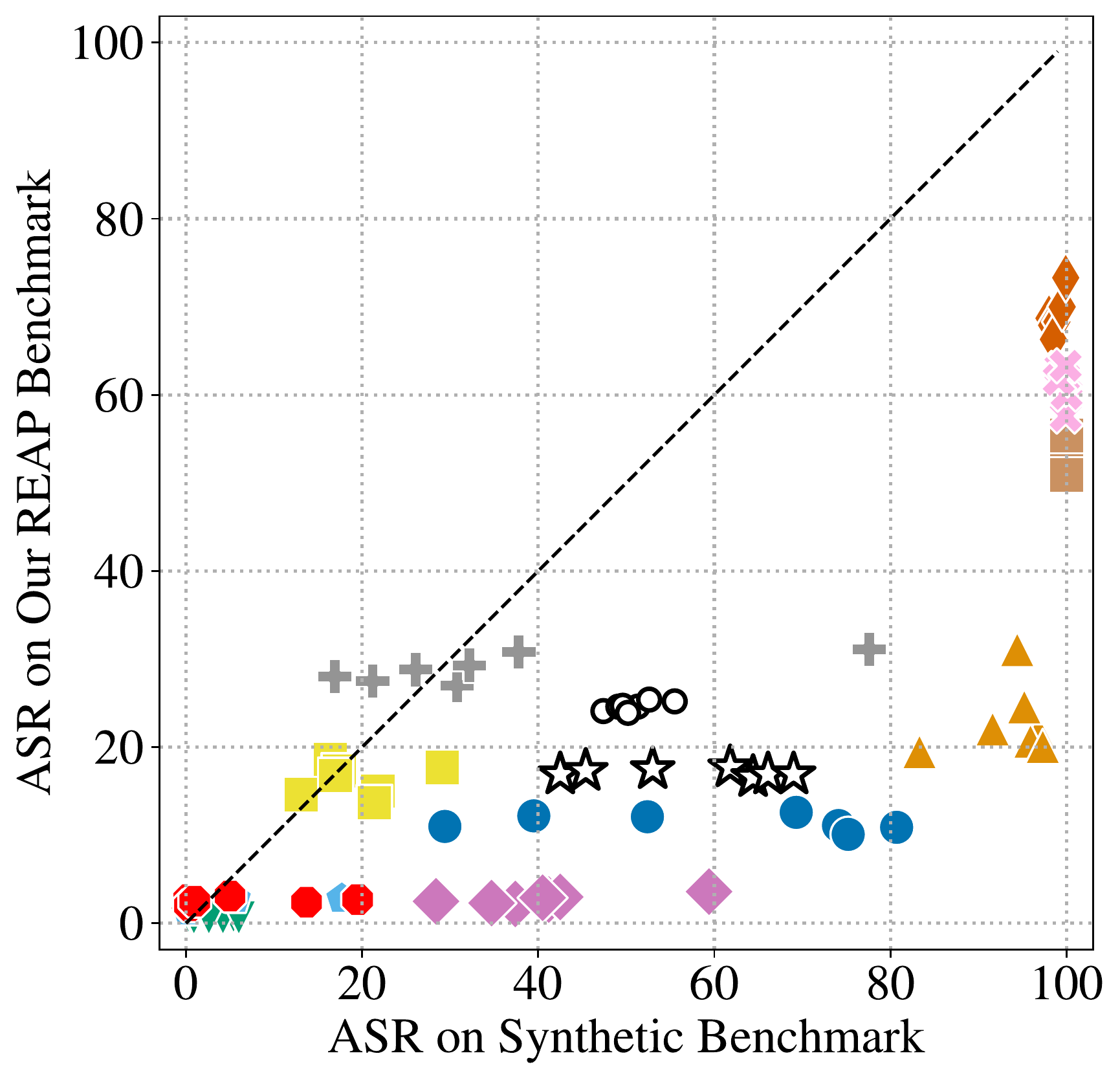}
      \caption{Rotation degrees}
      \label{fig:scatter_10x10_frcnn_rt}
   \end{subfigure}
   \hfill
   \begin{subfigure}[b]{0.32\textwidth}
      \centering
      \includegraphics[width=\textwidth]{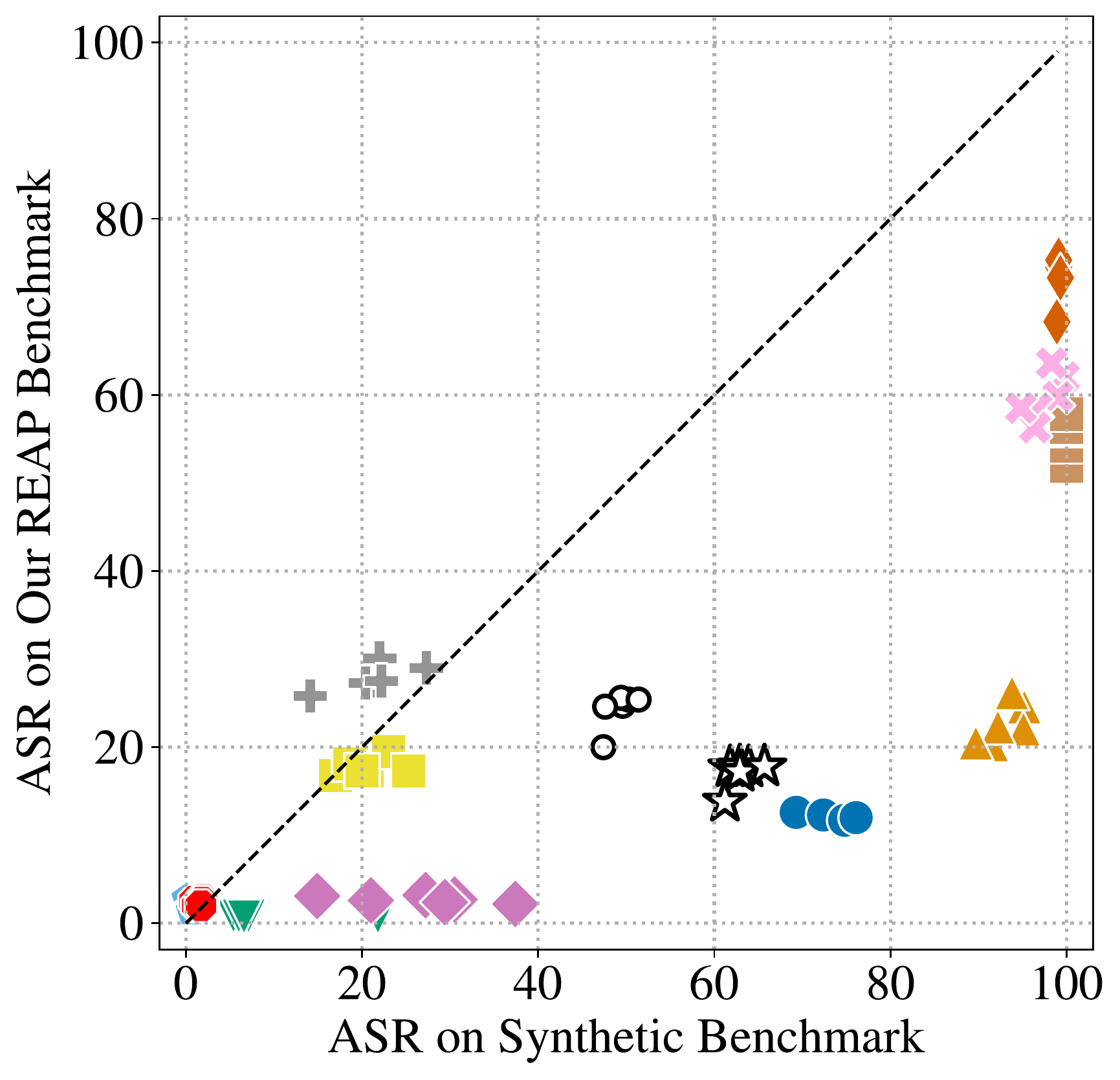}
      \caption{Color jitter intensity}
      \label{fig:scatter_10x10_frcnn_cj}
   \end{subfigure}
   \hfill~
   \caption{A scatter plot similar to \cref{fig:reap_vs_syn} where each point denotes a pair of ASRs measured by the synthetic and our \apbs{} benchmark for each class of the signs and also for each hyperparameter choice for the synthetic benchmark. Particularly, we sweep three hyperparameters that control (a) patch and object dimension, (b) the range of rotation degree used in EoT, and (c) the color jitter intensity used in EoT.}
   \label{fig:scatter}
\end{figure}

\begin{table}[t]
    \caption{ASR on different models under different attacks and optimizers. Higher ASR ($\uparrow$) means a stronger attack.}
    \label{tab:atk_opt}
    \vspace{-5pt}
    \small
    \centering
    \begin{tabular}{@{}llrrr@{}}
        \toprule
        Attacks                 & Optimizers & Faster R-CNN  & YOLOF         & DINO          \\ \midrule
        \multirow{2}{*}{RP2}    & Adam       & 39.0          & \textbf{48.0} & 31.2          \\
                                & PGD        & 37.9          & 45.4          & \textbf{31.6} \\ \cmidrule(l){2-5}
        \multirow{2}{*}{DPatch} & Adam       & \textbf{41.2} & 43.3          & 29.3          \\
                                & PGD        & 39.2          & 38.9          & 28.0          \\ \bottomrule
    \end{tabular}
\end{table}

\subsection{Attack Hyperparameters on the Synthetic Benchmark}

We conduct an additional ablation study to compare the effects of the hyperparameters of the synthetic benchmark as mentioned in \cref{ap:sec:exp_setup}.
\cref{fig:scatter} shows a similar scatter plot to \cref{fig:reap_vs_syn} but with all the hyperparameters we have swept.
This plot further strengthens the conclusions that the synthetic benchmark overestimates ASR and that it is not predictive of the ASR on our \apbs{} benchmark.
These observations persist across all the hyperparameter choices.

Additionally, \cref{fig:scatter} demonstrates that there is a large variation in the ASRs measured by the synthetic benchmark when the hyperparameters vary.
For instance, changing the rotation can affect the ASR up to 20\%--40\% for many signs.
This emphasizes that results reported on a synthetic benchmark are sensitive to its hyperparameters, and we should take special care when using one.
On the other hand, our \apbs{} benchmark does not have a similar set of hyperparameters to sweep over since the transformations as well as the sign sizes are fixed with respect to each image.

\begin{table}[t]
       \caption{The mAP scores ($\uparrow$) on our realistic benchmark when different choices of transformations are applied during evaluation. The default is using the perspective transform and the percentile (0.2) transform.}\label{tab:tf_ablation_full}
       \centering
       \small
       {\setlength\tabcolsep{4pt}
              \begin{tabular}{ll|rrrrrrrrrrr|r}
                     \toprule
                     Models & Transforms           & Circ & Tri  & UTri & Dia(S) & Dia(L) & Squ  & Rec(S) & Rec(M) & Rec(L) & Pen  & Oct  & \textit{Avg.} \\ \midrule
                     \multirowcell{10}[0pt][l]{Faster R-CNN}
                            & Default              & 48.0 & 54.9 & 62.4 & 53.3   & 61.0   & 41.1 & 24.9   & 52.4   & 59.0   & 60.1 & 67.5 & 53.2          \\
                     \cmidrule{3-14}
                            & Translate\&Scale     & 48.5 & 51.9 & 62.8 & 39.1   & 60.7   & 33.7 & 15.7   & 51.4   & 59.7   & 60.0 & 65.7 & 49.9          \\
                            & Affine               & 49.3 & 50.8 & 62.7 & 43.8   & 61.9   & 36.3 & 20.9   & 52.4   & 59.4   & 58.5 & 66.4 & 51.1          \\
                     \cmidrule{3-14}
                            & No Relight           & 27.3 & 40.9 & 59.6 & 25.8   & 54.3   & 26.3 & 6.7    & 46.1   & 39.0   & 52.5 & 53.6 & 39.3          \\
                            & Percentile (0.05)    & 44.3 & 49.8 & 61.9 & 33.7   & 57.9   & 34.3 & 14.4   & 51.2   & 55.6   & 56.3 & 61.5 & 47.4          \\
                            & Percentile (0.1)     & 44.6 & 51.0 & 61.5 & 33.1   & 60.1   & 34.4 & 17.1   & 52.4   & 57.5   & 56.9 & 60.1 & 48.1          \\
                            & Percentile (0.3)     & 49.6 & 52.8 & 63.0 & 51.5   & 62.0   & 40.0 & 25.3   & 54.9   & 60.4   & 60.4 & 66.5 & 53.3          \\
                            & Polynomial (HSV)     & 50.3 & 53.4 & 64.6 & 38.8   & 63.2   & 37.6 & 21.9   & 54.6   & 60.5   & 63.1 & 67.3 & 52.3          \\
                            & Polynomial (LAB)     & 49.4 & 54.9 & 67.4 & 45.1   & 64.9   & 38.6 & 11.5   & 53.2   & 46.4   & 57.5 & 61.6 & 50.0          \\
                            & Color Transfer (HSV) & 46.3 & 51.8 & 63.0 & 33.0   & 62.1   & 38.5 & 19.4   & 46.5   & 51.1   & 60.8 & 66.7 & 49.0          \\
                     \cmidrule{2-14}
                     \multirowcell{7}[0pt][l]{YOLOF}
                            & Default              & 42.9 & 42.2 & 63.2 & 32.8   & 55.8   & 40.7 & 5.2    & 52.3   & 58.6   & 61.1 & 66.3 & 47.4          \\
                     \cmidrule{3-14}
                            & Translate\&Scale     & 44.1 & 39.5 & 64.5 & 34.8   & 54.8   & 39.8 & 3.8    & 50.7   & 55.9   & 55.2 & 62.3 & 46.0          \\
                            & Affine               & 44.0 & 41.1 & 62.8 & 31.0   & 55.1   & 42.2 & 5.0    & 52.6   & 59.1   & 59.5 & 66.4 & 47.2          \\
                     \cmidrule{3-14}
                            & No Relight           & 26.5 & 21.2 & 56.0 & 17.3   & 46.7   & 33.4 & 0.5    & 12.7   & 28.6   & 47.1 & 46.4 & 30.6          \\
                            & Percentile (0.05)    & 40.3 & 41.2 & 61.0 & 25.9   & 52.2   & 39.0 & 1.6    & 36.9   & 48.4   & 50.6 & 62.0 & 41.7          \\
                            & Percentile (0.1)     & 42.0 & 41.4 & 61.2 & 28.7   & 53.0   & 39.6 & 2.8    & 43.4   & 54.3   & 58.6 & 62.6 & 44.3          \\
                            & Percentile (0.3)     & 45.4 & 44.1 & 65.1 & 39.8   & 56.9   & 41.1 & 11.7   & 56.9   & 60.8   & 64.6 & 68.0 & 50.4          \\
                     \bottomrule
              \end{tabular}
       }
\end{table}

\subsection{Detailed Effects of the Transform Methods}

Here, we report the full results by class of \cref{fig:transforms_mAP} in \cref{tab:tf_ablation_full}.
The relighting transform still affects the effectiveness of the patch attack to a greater degree than the geometric transform.
This trend is consistent not only on average but also across all the sign classes.

\subsection{Per-Instance Attack}\label{ap:ssec:per_instance}

As another ablation study, we experiment with the worst-case possible attack on our benchmark where the adversary can generate a unique adversarial patch for each image and is also aware of how the patch will appear in the image exactly.
This setting is similar to the commonly studied ``white-box'' attack in the adversarial example literature.
This threat model is particularly unrealistic for patch attacks because, in the real world, the adversary cannot predict apriori how the video or the image of the patch will be taken.
Nonetheless, theoretically, this measurement is useful because the ASR in this setting should be the upper bound of any other setting including the ``per-class'' threat model we have considered throughout the paper.

\cref{tab:transfer} report the per-instance ASR on our \apbs{} benchmark.
We only compute the ASR for Faster R-CNN because this experiment is computationally expensive even when we reduce the attack iterations from 1,000 to 100.
This experiment takes about 7 days to finish on an Nvidia GTX V100 GPU.
The per-instance attack results in about 10 percentage points higher ASR than the per-class attack on Adv. Faster R-CNN and Adv. YOLOF but only 2.6 percentage points on Adv. DINO.
Regardless, this increase in the ASR is huge in the relative sense.

There are two ways to interpret this observation: first, it could mean that an object detection model may be more robust to physical attacks than the researchers expect, and this makes coming up with an effective defense easier.
The second way to view this result is that the previously proposed attack algorithms are far from optimal, and there is a large room for improvement from the attacker's side.

\begin{table}[t]
    \caption{Results on the adversarial patch that covers the entire signs (\psize{50}{50}). Here, we use DPatch attack with PGD optimizer. Per-class metrics for Adv. DINO model is shown in \cref{tab:adv_dino_cover_all}.}
    \label{tab:cover_all}
    \small
    \centering
    \begin{tabular}{@{}lrrrrrr@{}}
        \toprule
        \multirowcell{2}[0pt][l]{Metrics} & \multicolumn{6}{c}{Models}                                                             \\ \cmidrule(lr){2-7}
                                          & Faster R-CNN               & YOLOF & DINO & Adv. Faster R-CNN & Adv. YOLOF & Adv. DINO \\ \midrule
        FNR ($\downarrow$)                & 99.7                       & 99.9  & 99.8 & 99.3              & 99.3       & 66.0      \\
        ASR ($\downarrow$)                & 99.7                       & 99.8  & 99.8 & 99.2              & 99.1       & 66.2      \\
        mAP ($\uparrow$)                  & 0.3                        & 0.2   & 0.5  & 0.9               & 1.2        & 16.8      \\ \bottomrule
    \end{tabular}
\end{table}
\begin{table}[t]
    \centering
    \small
    \caption{Results on the adversarial patch that covers the entire signs (\psize{50}{50}) on Adv. DINO model. RP2 attack performs even worse than DPatch here.}
    \label{tab:adv_dino_cover_all}
    \begin{tabular}{llrrrrrrrrrrrr}
        \toprule
        Attacks & Metrics & Circ & Tri  & UTri & Dia(S) & Dia(L) & Squ  & Rec(S) & Rec(M) & Rec(L) & Pen  & Oct  & \textit{Avg} \\ \midrule
        \multirowcell{3}[0pt][l]{DPatch}
                & FNR     & 43.6 & 58.9 & 97.6 & 1.9    & 94.6   & 62.1 & 9.0    & 82.8   & 100.0  & 92.2 & 83.2 & 66.0         \\
                & ASR     & 46.7 & 57.5 & 97.6 & 2.0    & 94.5   & 63.9 & 7.8    & 83.1   & 100.0  & 92.1 & 83.3 & 66.2         \\
                & mAP     & 9.6  & 33.9 & 27.1 & 60.1   & 8.5    & 8.1  & 25.6   & 3.9    & 3.1    & 3.2  & 2.4  & 16.8         \\ \cmidrule(lr){2-14}
        \multirowcell{3}[0pt][l]{RP2}
                & FNR     & 9.9  & 55.8 & 97.5 & 1.2    & 93.8   & 7.2  & 7.9    & 77.4   & 100.0  & 93.5 & 83.2 & 57.0         \\
                & ASR     & 9.4  & 54.4 & 97.5 & 1.2    & 93.7   & 7.4  & 6.3    & 77.1   & 100.0  & 93.4 & 83.1 & 56.7         \\
                & mAP     & 18.7 & 35.6 & 28.6 & 66.6   & 14.9   & 19.8 & 29.7   & 4.9    & 3.8    & 8.1  & 9.2  & 21.8         \\ \bottomrule
    \end{tabular}
\end{table}

\subsection{Patch Attack that Covers the Entire Signs}\label{ap:ssec:cover_all}

Trying to unveil the reason behind the robustness of Adv. DINO on \apbs{}, we want to test how it would perform in an extreme case where the adversarial patch may completely cover up the entire sign.
We choose a patch size of \psize{50}{50} which is larger than all types of the signs and centered it in the middle.

Surprisingly, the ASR for some classes still does not reach 100\%: the average is 66\% (\cref{tab:cover_all}).
However, the class-wise results in \cref{tab:adv_dino_cover_all} show that the model is completely confused among the classes that have the same shape but only differ by sizes, e.g., diamond (S/L) and rectangle (S/M/L).
One explanation is that the model learns a strong shape bias in detecting the signs, and hence, it still often detects the signs correctly based on the shape alone.
The other possibility is the existing attacks find a suboptimal patch yielding a false sense of security.
This outcome is, however, specific to Adv. DINO as the other models, including a normally trained DINO, do get fooled $>$99\% of the times under this patch size (\cref{tab:cover_all}).

With \apb{} benchmark, the shape information alone becomes less useful to the models.
When using the patch size that covers the whole sign, we find that Adv. DINO is now no longer robust (78\% ASR) as the shape information alone is not sufficient to distinguish all the signs.

\clearpage

\begin{figure}[t]
   \centering
   \includegraphics[width=0.9\textwidth]{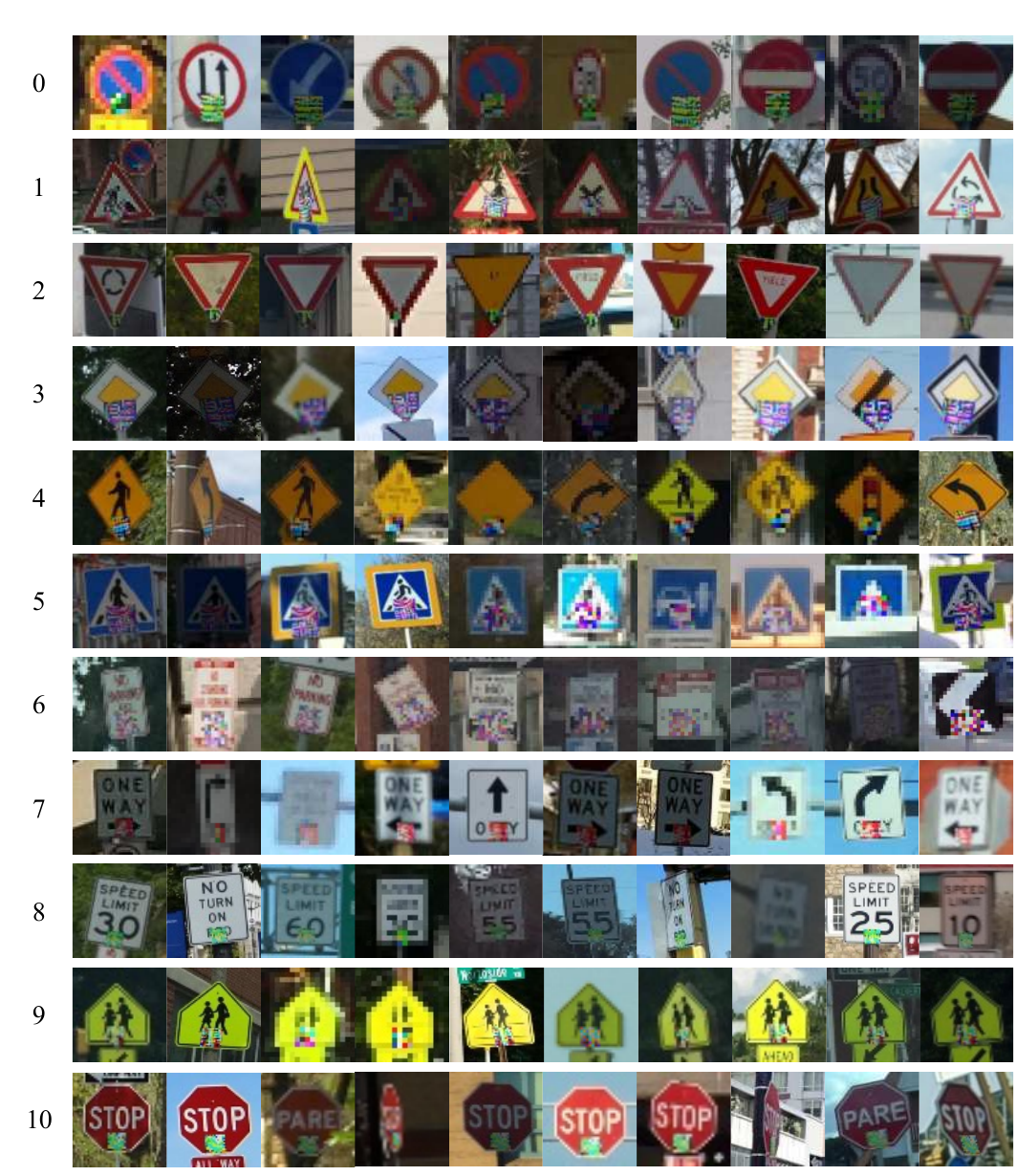}
   \caption{Random samples of traffic signs from each of the 11 classes (one per row) applied with a \psize{10}{10} adversarial patch using the transforms from our \apbs{} benchmark. The numbers on the left indicate class ID (0: Circle, 1: Triangle, 2: Upside-down triangle, 3: Diamond (S), 4: Diamond (L), 5: Square, 6: Rectangle (S), 7: Rectangle (M), 8: Rectangle (L), 9: Pentagon, 10: Octagon).}
   \label{fig:sign_samples}
\end{figure}

\begin{figure}[t]
   \centering
   \begin{subfigure}[b]{0.49\textwidth}
      \centering
      \includegraphics[width=\textwidth]{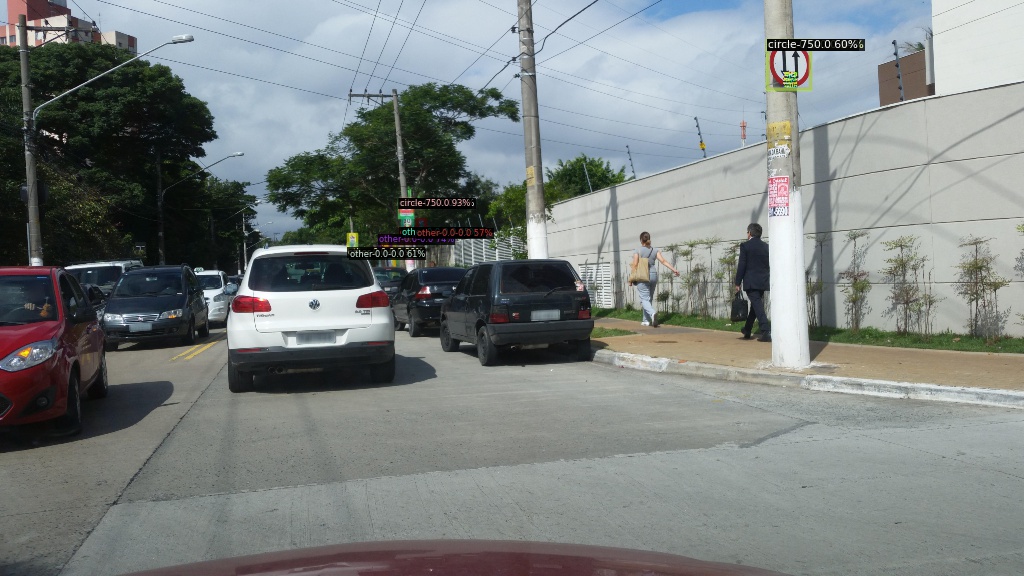}
      \caption{Circle}
   \end{subfigure}
   \hfill
   \begin{subfigure}[b]{0.49\textwidth}
      \centering
      \includegraphics[width=\textwidth]{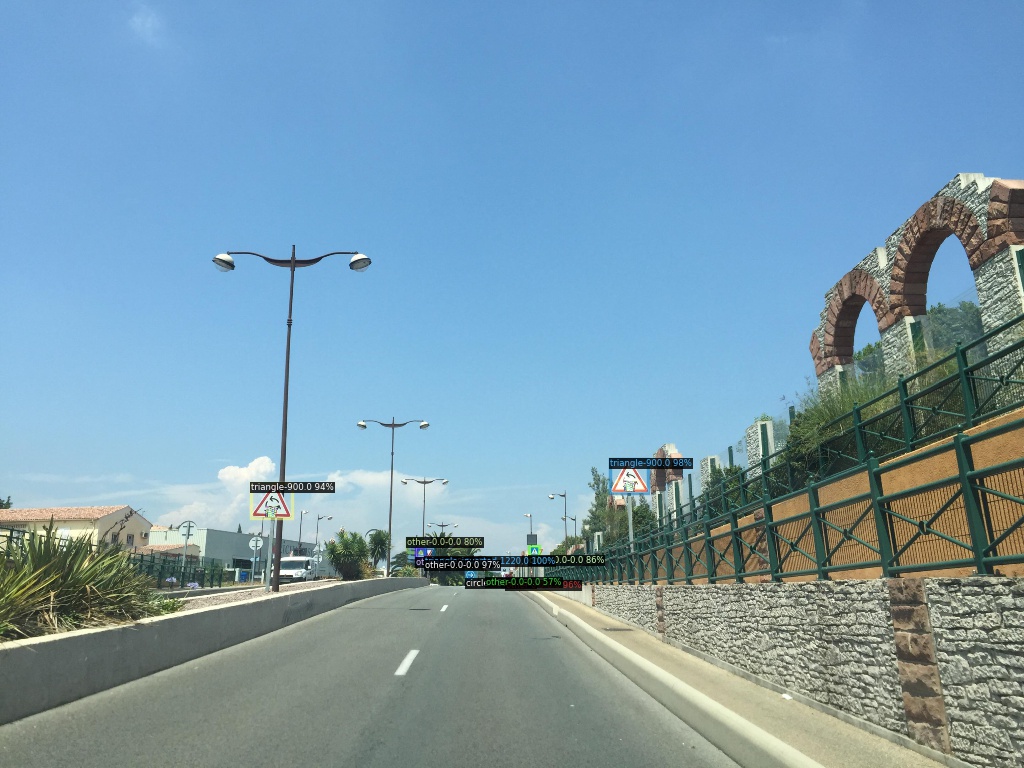}
      \caption{Triangle}
   \end{subfigure}
   \hfill
   \begin{subfigure}[b]{0.49\textwidth}
      \centering
      \includegraphics[width=\textwidth]{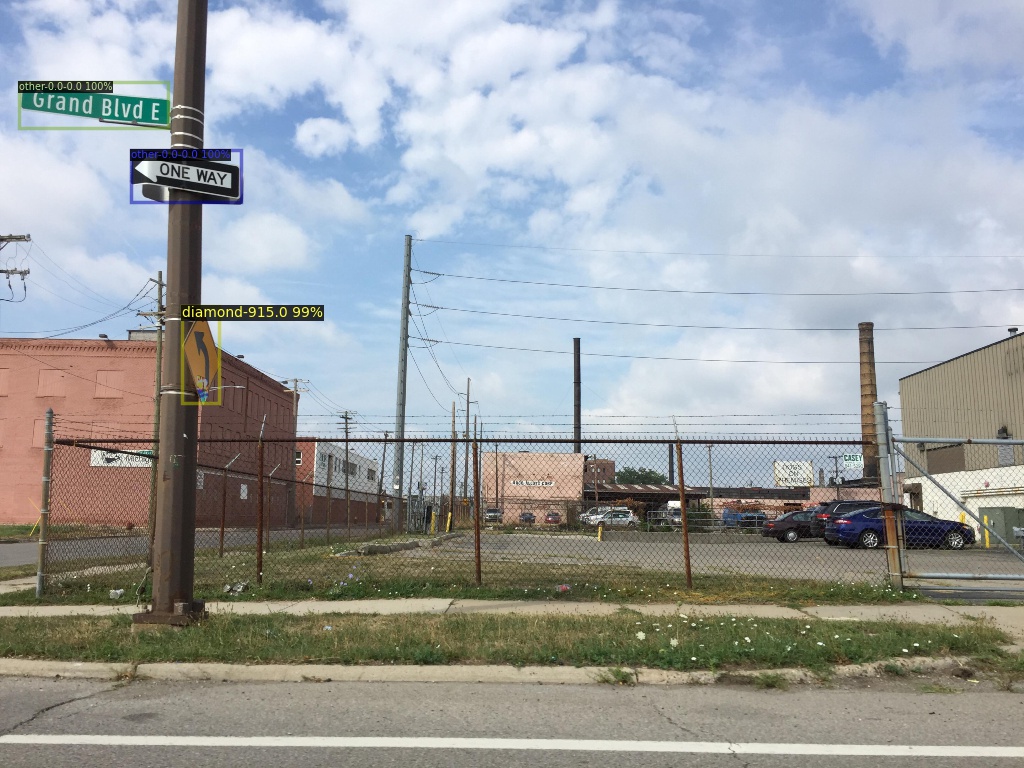}
      \caption{Diamond (L)}
   \end{subfigure}
   \hfill
   \begin{subfigure}[b]{0.49\textwidth}
      \centering
      \includegraphics[width=\textwidth]{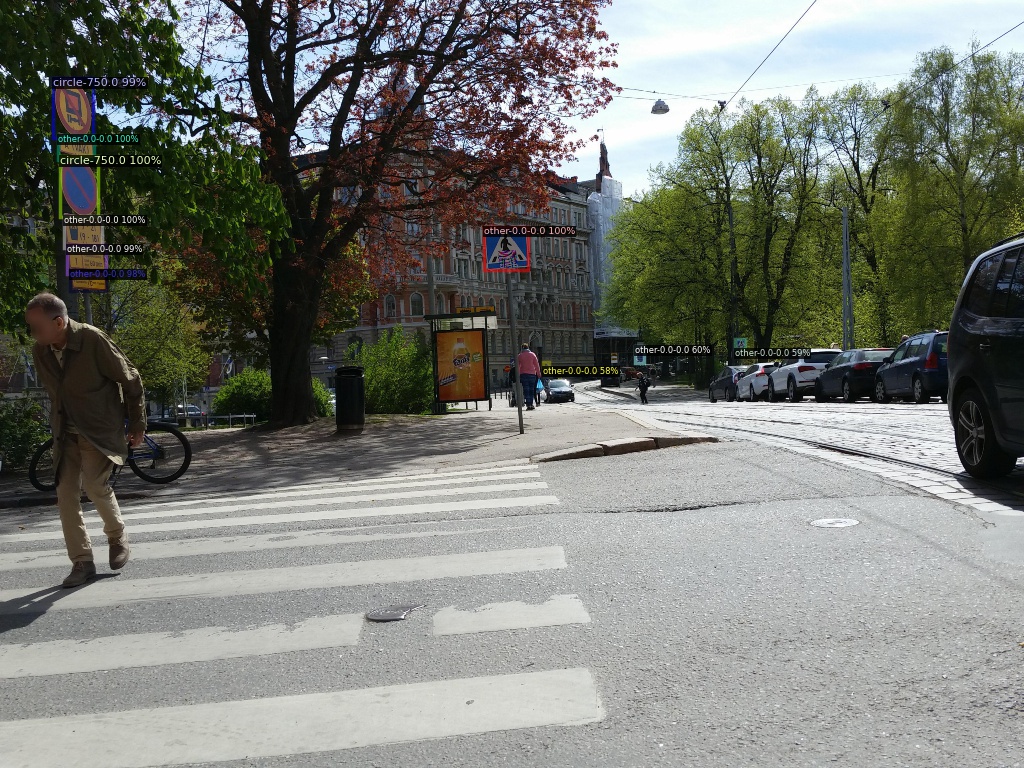}
      \caption{Square}
   \end{subfigure}
   \hfill
   \begin{subfigure}[b]{0.49\textwidth}
      \centering
      \includegraphics[width=\textwidth]{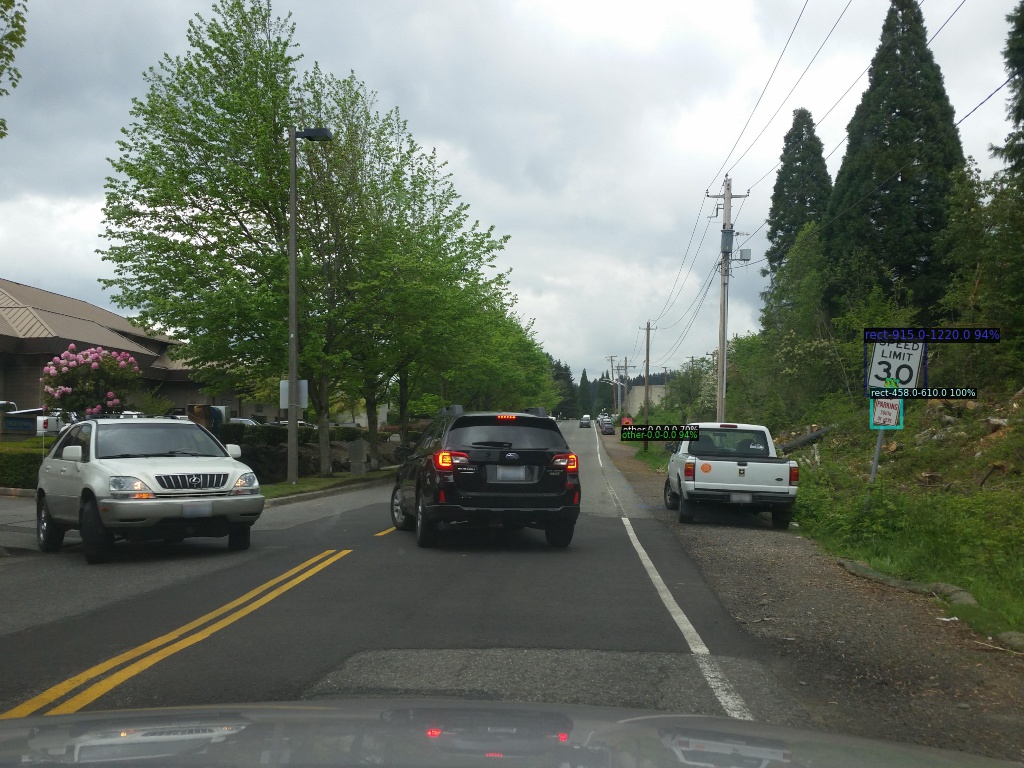}
      \caption{Rectangle (L)}
   \end{subfigure}
   \hfill
   \begin{subfigure}[b]{0.49\textwidth}
      \centering
      \includegraphics[width=\textwidth]{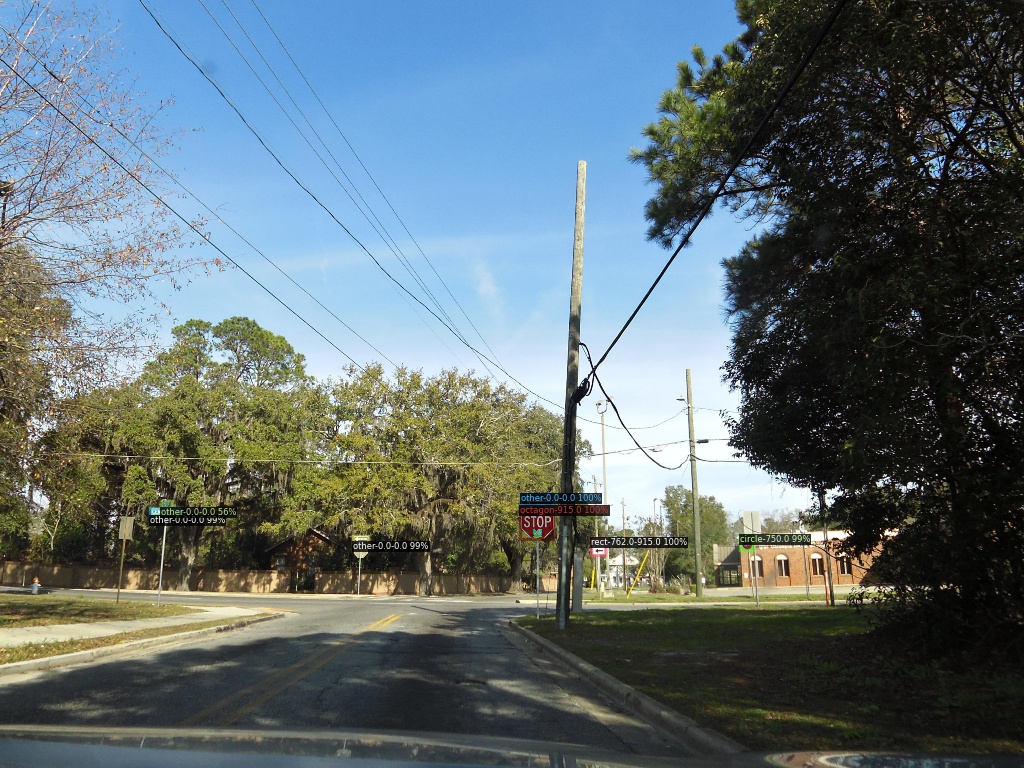}
      \caption{Octagon}
   \end{subfigure}
   \caption{Examples of images from our benchmark after applying the \inch{10}$\times$\inch{10} patch. The sub-caption indicates the target sign class. We try to select images that the signs are large enough to see on the printed paper.}
   \label{fig:visual_10x10}
\end{figure}

\section{Additional Visualization of the Benchmark} \label{ap:sec:more_viz}

In \cref{fig:visual_10x10}, we select six images from our benchmark with the \inch{10}$\times$\inch{10} patch applied, one from each sign class.

\end{document}